\documentclass[screen,authorversion,nonacm,review=false,timestamp=false]{acmart}

\usepackage{booktabs}            % professional-quality tables
\usepackage{multirow}    
\usepackage{float}% tabular cells spanning multiple rows
\usepackage{amsfonts} 
\usepackage{subcaption}% blackboard math symbols
\usepackage{graphicx} % figures
\usepackage{duckuments}          % sample images
\usepackage{subcaption}          % fig a and b labels

\newcommand{\modelparentdist}{\Tilde{\mathcal{P}}}
\newcommand{\trueparentdist}{\mathcal{P}}
\usepackage{caption}
\usepackage{algpseudocode}
\usepackage{algorithm}
\usepackage{amsthm}              % theorem environments
\newtheorem{theorem}{Theorem}[section]
\newtheorem*{theorem*}{Theorem}
\theoremstyle{definition}

\newtheorem*{idea}{Idea}

\AtBeginDocument{%
  }

\setcopyright{acmlicensed}
\copyrightyear{2025}
\acmYear{2025}
\acmDOI{XXXXXXX.XXXXXXX}
\acmConference[KDD '25]{Machine Learning on Graphs in the Era of Generative Artificial Intelligence}{August 03--07,2025}{Toronto, Canada}
\begin{document}

\title[Neural Algorithmic Reasoning with Multiple Correct Solutions]{Neural Algorithmic Reasoning with Multiple Correct Solutions}

\author{Zeno Kujawa}
\authornote{Both authors contributed equally to this research.}
\affiliation{%
  \institution{Northeastern University}
  \city{Boston}
  \state{Massachusetts}
  \country{USA}
}
\email{kujawa.z@northeastern.edu}

\author{John Poole}
\authornotemark[1]
\affiliation{%
  \institution{University of Cambridge}
  \city{Cambridge}
  \country{United Kingdom}}
\email{jmp251@cam.ac.uk}

\author{Dobrik Georgiev}
\affiliation{%
  \institution{University of Cambridge}
  \city{Cambridge}
  \country{United Kingdom}}
  \email{dgg30@cam.ac.uk}

\author{Danilo Numeroso}
\affiliation{%
 \institution{Università di Pisa}
 \city{Pisa}
 \country{Italy}}
 \email{danilo.numeroso@phd.unipi.it}

\author{Henry Fleischmann}
\affiliation{%
  \institution{Carnegie Mellon University}
  \city{Pittsburgh}
  \state{Pennsylvania}
  \country{USA}}
  \email{hfleisch@andrew.cmu.edu}

\author{Pietro Liò}
\affiliation{%
  \institution{University of Cambridge}
  \city{Cambridge}
  \country{United Kingdom}}
  \email{Pietro.Lio@cam.ac.uk}

\renewcommand{\shortauthors}{Kujawa et al.}

\begin{abstract} %Optimization problems may have multiple correct solutions. %Classical algorithms may produce multiple correct solutions.
Neural Algorithmic Reasoning (NAR) extends classical algorithms to higher dimensional data. However, canonical implementations of NAR train neural networks to return only a single solution, even when there are multiple correct solutions to a problem, such as single-source shortest paths. For some applications, it is desirable to recover more than one correct solution. To that end, we give the first method for NAR with multiple solutions. We demonstrate our method on two classical algorithms: Bellman-Ford (BF) and Depth-First Search (DFS), favouring deeper insight into two algorithms over a broader survey of algorithms. This method involves generating appropriate training data as well as sampling and validating solutions from model output. Each step of our method, which can serve as a framework for neural algorithmic reasoning beyond the tasks presented in this paper, might be of independent interest to the field and our results represent the first attempt at this task in the NAR literature. 
\end{abstract}



\received{20 February 2007}
\received[revised]{12 March 2009}
\received[accepted]{5 June 2009}

\maketitle
\section{Introduction}
%\subsection{From Classical Algorithms to Neural Algorithmic Reasoning \cite{nar-blueprint}} % algorithmic guarantees apply only to the abstract inputs; so algorithmic guarantees are only as good as the transformation from real inputs to the abstract input
Classical algorithms like Merge Sort guarantee correctness and generalize perfectly on abstract inputs. Unfortunately, real problems are rarely abstract inputs. Consider pathfinding: to give traffic directions via a shortest paths algorithm, one must compress physical distance and congestion data for each street into single-number edge labels, thereby losing information. Bad edge labels mean bad directions, even despite a good algorithm. Neural Networks (NNs) avoid this problem by directly operating on high-dimensional data. Instead of compressing distance and congestion to a single path-length edge label, an NN can operate directly on distance and congestion data. The chief hurdle to NN performance is failure to generalize: common architectures are fragile to changes in input size and structure — exactly where algorithms succeed. It is thus natural to wonder whether an NN might learn to generalize from an algorithm. To that end, Neural Algorithmic Reasoning is the field that aims to train NNs to mimic the running of classical algorithms \cite{nar-blueprint}. A benefit of this approach is staying high-dimensional: no information is lost when projecting high-dimensional real data into a single abstract output for a classical algorithm \cite{nar-blueprint}. Thus far, NNs have been trained to mimic algorithms which recover \emph{one} solution from the space of possible correct solutions \cite{clrs-benchmark, ibarz2022generalist, hint-reversal, no-hints, conar, veličković2020neuralexecutiongraphalgorithms,bohde2024on}. Ordinarily, an NN trained to perform DFS produces an array, which one can view as a distribution containing the probabilities of different solutions, from which the likeliest solution is returned as the model output. We adapt the model input to reflect multiple correct solutions and present methods of retrieving multiple solutions from the model output distribution.%\footnote{An anonymized repository of the code used in the vast majority of experiments can be found \underline{\href{https://anonymous.4open.science/r/NAR_Multiple_Solutions-2231}{here}}. We plan to release a full, de-anonymized repository for the camera-ready submission.} 
Using only one solution in NAR may be vulnerable to local minima in the model training. By diversifying the space of solutions returned by NAR, we may decrease the risk of solutions originating from a local minimum. For applications like cyberphysical systems, the flexibility of multiple solutions is essential. A single solution, for reasons of cost, safety, or compliance, may not be amenable to a final task.

\subsection{Contributions}
We give the first analysis of Neural Algorithmic Reasoners which return more than one solution. For training examples, we create distributions of solutions from multiple runs of classical algorithms with randomized tie-breaking. We train NNs to predict the generated distribution of solutions; and we leverage randomness in algorithm-specific ways to extract distinct solutions from the NN-predicted distributions.
We also discuss methods of evaluating the solutions obtained through our methods. Thus we offer three contributions that might each be of independent interest: (1) a method for training a neural network on a distribution of correct solutions, (2) a method for extracting multiple solutions from the distribution the NN predicts, (3) an evaluation of the diversity and correctness of solutions our method yields. Together, these three contributions can serve as a framework for further research into Neural Algorithmic Reasoning with multiple solutions for each problem instance. 
\section{Methodology}
\begin{figure}[t]
    \centering
    \includegraphics[width=\linewidth]{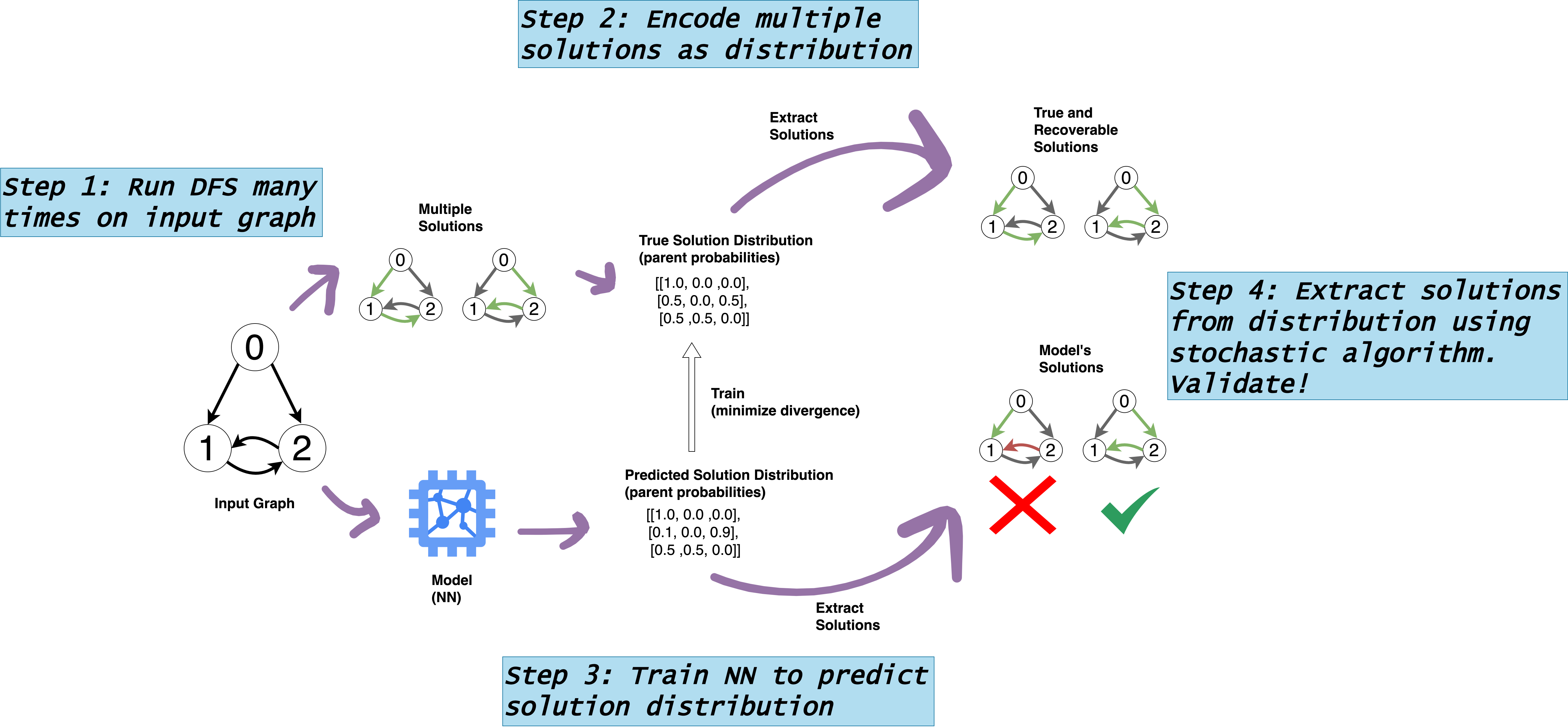}
    \caption{Blueprint for NAR with multiple correct solutions. For Step 2 see Figure \ref{fig:parent-dist}. For Step 4 see Figure \ref{fig:extract-solutions} and Appendices \ref{sec: Appendix: Randomness in Sampling (DFS)} \ref{sec: Appendix: Randomness in Sampling (BF)}. Best viewed on a screen.}
    \label{fig:multiple-sol-graphical-abstract}
\end{figure}

To produce multiple solutions for a single graph, we train a NN to predict a distribution of solutions for each individual graph (solutions being individual DFS trees or BF paths, their distribution represented as distributions over child-parent pairs, as discussed in Appendices \ref{sec: Appendix Randomness in training (DFS)} and \ref{sec: Appendix Randomness in training (BF)}). Ordinarily, the CLRS benchmark \cite{clrs-benchmark} generates training data by running classical algorithms with deterministic tie-breaking. For example, the input graph in Figure \ref{fig:multiple-sol-graphical-abstract} has two valid orders of depth-first traversal, but classical DFS will only return one based on which edge it explores first. To create our training examples, we run a classical algorithm (e.g. DFS) multiple times with randomized tiebreaking, and average the outputs (e.g. individual DFS trees) into a distribution of solutions.  As we want the NN to predict a distribution, we train to minimize a distribution divergence (Kullback-Leibler: KL) \cite{kldiv_1951} between the input distribution of solution and that given by the NN. As our final goal is to recover multiple solutions for an input graph, we develop algorithm-specific stochastic sampling methods. Thus, there are two sources of randomness in our process: the translation from a deterministic algorithm to a probability distribution over predecessor-successor relationships uses randomness, as does the sampling of predecessor arrays from the model output distribution.
Stochasticity of sampling methods means samples are likely to have multiple different solutions. For DFS, we sample the immediate parents of vertices until we reach the root vertex (which can be seen as executing DFS in reverse on a graph); our two DFS extractors, which we call \textit{Upwards} and \textit{AltUpwards}, differ in the manner in which they discard potential solutions (Appendix \ref{sec: Appendix: Randomness in Sampling (DFS)}). For BF, \textit{Beam} samples a number of parents for non-source vertices according to parent-probabilities and chooses the lowest cost path arising (Appendix \ref{sec: Appendix: Randomness in Sampling (BF)}). Also for BF, \textit{Greedy} samples parents for each non-source vertex and chooses the parent with the lowest-cost outgoing edge for each non-source vertex (Appendix \ref{sec: Appendix: Randomness in Sampling (BF)}).

\begin{figure}[t]
    \centering
    \includegraphics[width=0.8\linewidth]{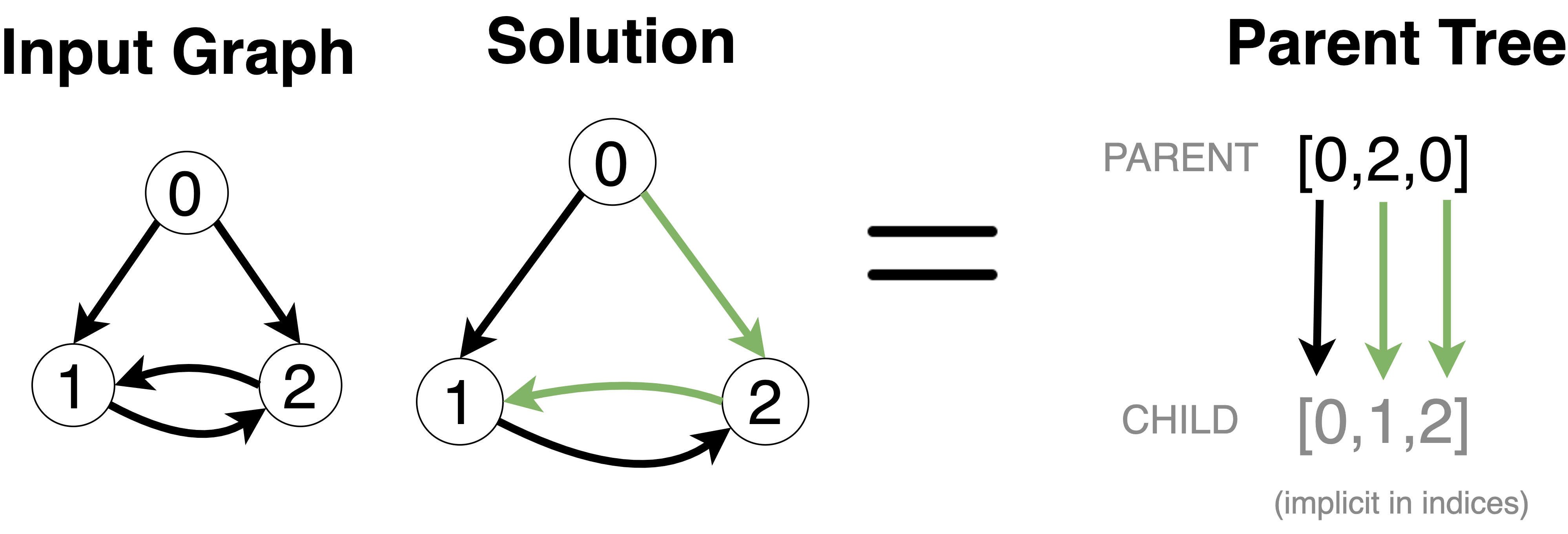}
    \caption{Parent Tree encoding of a single DFS solution. The start node, 0, is its own parent, represented by the value 0 at index 0. The value 2 at index 1 indicates that vertex 2 is the parent of vertex 1. Similarly, the value 0 at index 2 indicates that vertex 0 is the parent of vertex 2.} %Henceforth we draw only the parent list: the children are implicit in the indices.
    \label{fig:enter-label}
\end{figure}

\begin{figure}[t]
    \centering
    \includegraphics[width=0.8\linewidth]{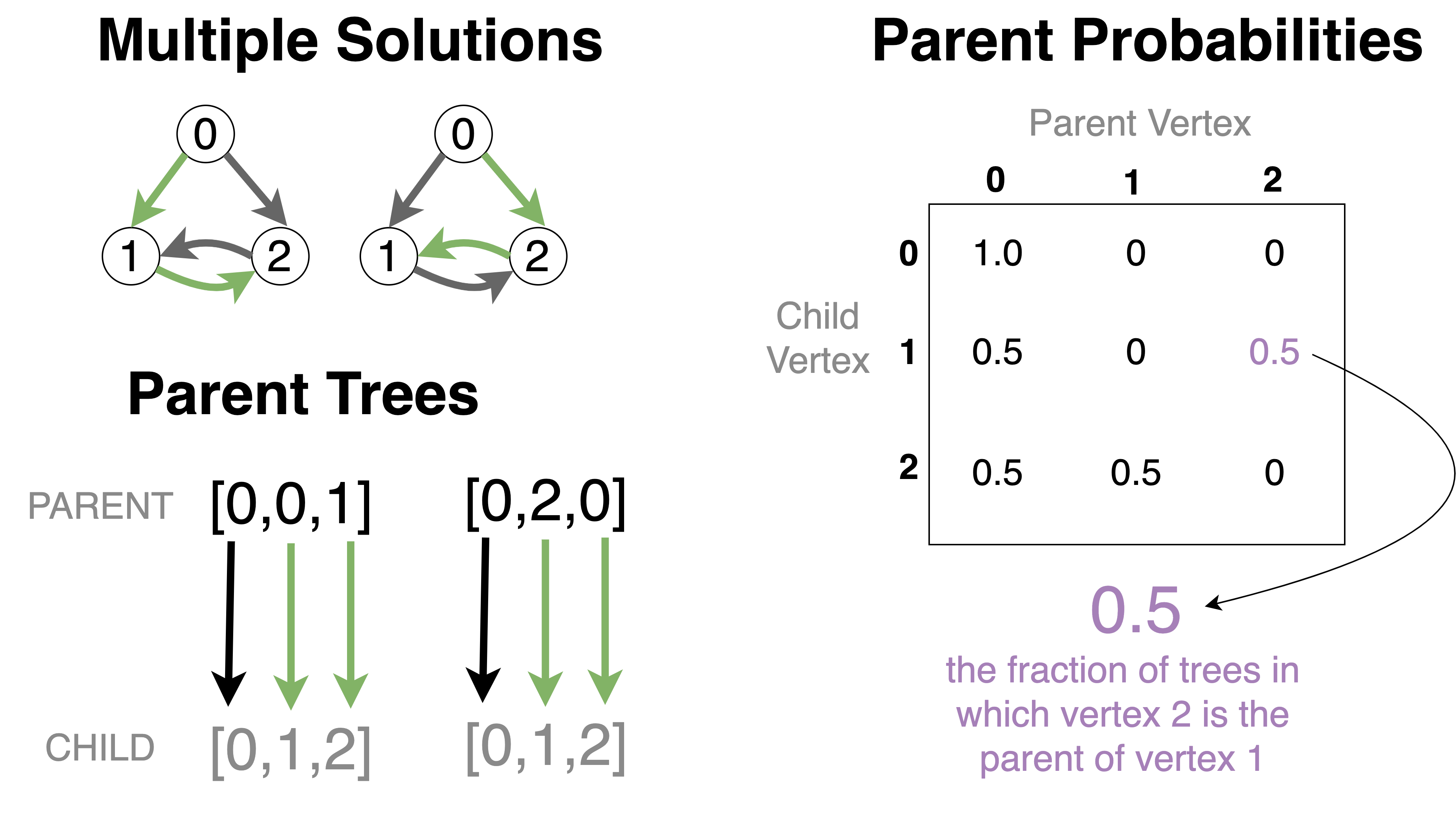}
    \caption{Parent Probability Distribution encoding of multiple DFS solutions. The value 0.5 in row $1$ column $2$ indicates that vertex 2 is a parent of vertex 1 in 50\% of correct parent trees. We call the parent probabilities obtained by repeatedly running an algorithm the \textit{true parent distribution} $\trueparentdist$. The model outputs, which we aim to be as close as possible to $\trueparentdist$, are denoted as $\modelparentdist$.} \label{fig:parent-dist}
\end{figure}
\begin{figure}[t]
    \centering
    \includegraphics[width=\linewidth]{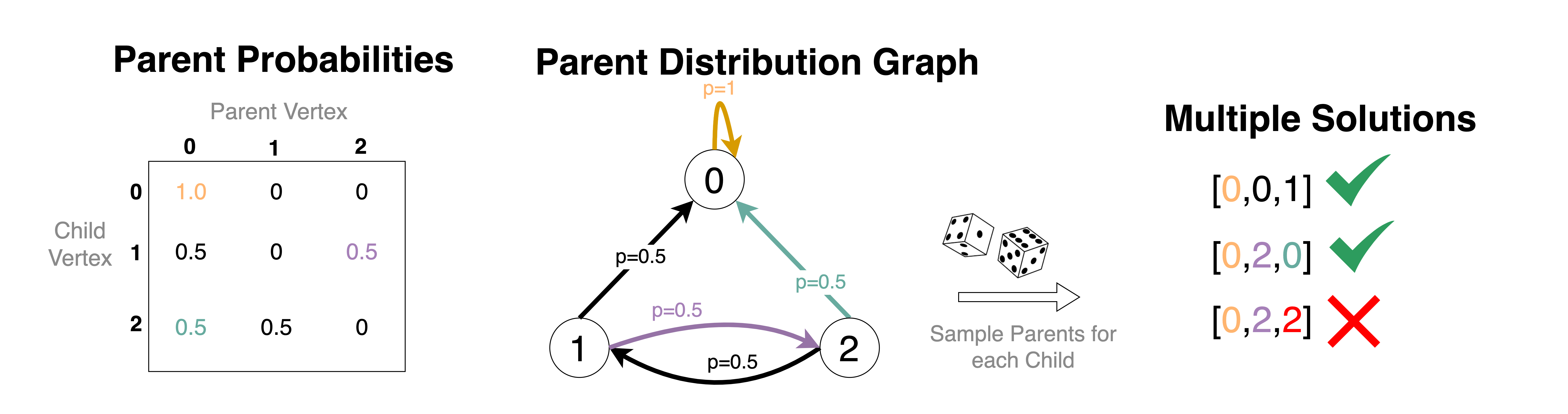}
    \caption{DFS tree extraction from a parent probability matrix:  $\modelparentdist$ or $\trueparentdist$. The parent tree [0,2,2] is incorrect because there is no solution in which 2 is its own parent. [0,0,0] is incorrect, despite there being solutions in which 0 is the parent of each vertex, because there is no solution in which 0 is the parent of every vertex: if 0 is the parent of 1, 1 must be the parent of 2. Best viewed on a screen.}
    \label{fig:extract-solutions}
    %\zeno{would cut this: Extracting Multiple Parent Trees from a Distribution of Parent Probabilities. The matrix of parent probabilities can be seen as an adjacency matrix for a weighted digraph where edges point from child to parent.} The process from a parent probability matrix $\trueparentdist$ or $\modelparentdist$  to a DFS tree. \zeno{would cut this: Walking with randomness across the graph from child to parent produces a solution, in this case a parent tree.} The correctness of a parent tree depends on the original input graph and the specific algorithm \zeno{would cut: (a dfs parent tree has different conditions for correctness than a single-source shortest paths parent tree)}. In the case of DFS, [0,2,2] is invalid since there is no solution in the solution distribution with an edge from 2 to itself, but [0,0,0] would also be incorrect, even though all nodes have edges to 0, as DFS forbids backtracking to 0 until paths outgoing from 1 have been explored. This provides an example of why validation of our sampled trees is critical.}
\end{figure}
% %\footnote{The distribution of parent probabilities, equally called the solution distribution, can also be thought-of as a weighted digraph, where an edge $(s,t)$ with weight 0.5 indicates a 50\% probability that vertex $t$ is the parent of vertex $s$.} %\footnote{Note $\trueparentdist$ is an approximation.}

\subsection{DFS}
We consider a generalization of the deterministic DFS setting implemented in the CLRS benchmark \cite{clrs-benchmark}. In our setting, edge-exploration is random: from a given vertex, any of the outgoing edges may be explored first. To decrease the computational burden of verifying solutions, we require that once DFS backtracks past the root of a connected component, it restarts its search at the lowest-indexed unvisited vertex. We view this generalization is natural, modelling a setting where one conducts DFS according to a chosen sequence of starting points, without assumptions about which edges to pursue first.

%Broadening to a DFS setting where any outgoing edge might be explored first allows us to consider multiple solutions correct. \john{We now say this in figure captions, right? Figure \ref{fig:dfs_multpath} gives a simple example of a situation where multiple correct DFS-trees exist when randomising the order of exploration while keeping the start node fixed.} Keeping the property that connected components will be explored according to a linear order helps us determine the correctness of a proposed solution along necessary conditions we introduce. We also think it has the advantage of naturality, modelling a setting where one conducts DFS along connected components according to an inductive bias of starting points.

\begin{figure}
    \centering
    \includegraphics[width=0.4\linewidth]{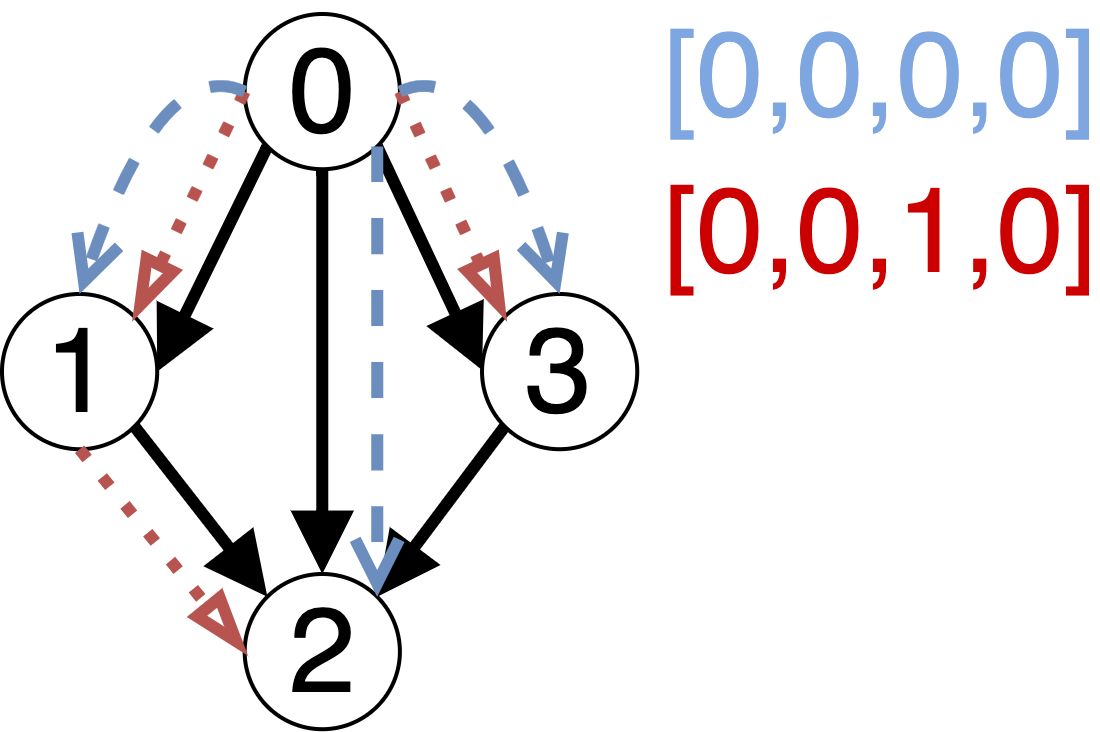}
    \caption{A graph with multiple correct DFS solutions. Graph edges are drawn in black. The red tree would be the result if vertices were explored in alphabetical order, while the blue tree is one of multiple other posible valid DFS trees. Note also how the DFS-tree is encoded as a predecessor array (with the entry at position $i$ is the predecessor of the $i^{th}$ vertex in alphabetical order, which we henceforth call $\pi$.}
    \label{fig:dfs_multpath}
\end{figure}

\subsubsection{Randomness in training}%\label{sec: Methodology Randomness in training (DFS)}
We generate an empirical parent distribution $\trueparentdist$ for each graph in the test set (Figure \ref{fig:parent-dist}) by running our randomized version of DFS 20 times (we conclude from our results in Appendix \ref{sec: test re-run numbers} that 20 runs provide a good approximation of the target distribution for each graph). We then train the model to minimize the KL-divergence between its predicted distribution $\modelparentdist$ and $\trueparentdist$ (Appendix \ref{sec: Appendix Randomness in training (DFS)}). %We obtain $\trueparentdist$ by combining 20 solutions generated by a DFS algorithm that randomly tiebreaks between outgoing edges. We choose 20 solutions for efficiency (Appendix \ref{sec: test re-run numbers}). We compute the fraction of solutions in which node $s$ is the parent of node $t$, which we say is the parent probability, as shown in Figure \ref{fig:parent-dist}. Algorithm \ref{alg:dfsOrdered} gives detailed description of our particular DFS algorithm, with the line that randomizes edge-exploration highlighted in red.

\subsubsection{Randomness in sampling}%\label{sec: Methodology: Randomness in Sampling (DFS)}
%To address the aim of the project, it is critical to solve the problem of obtaining valid DFS solutions from the distribution $\modelparentdist$ of parents. This section concerns itself with sampling methods, while the next section describes some necessary conditions against which sampled solutions are checked. The first method we employ is not strictly sampling, as we simply assume the parent of each node $x$ to be the node $u$ such that:

We compare 4 methods, which we refer to as \textit{Upwards altUpwards, Argmax}, and \textit{Random}, for extracting DFS solutions from a parent distribution, $\modelparentdist$. Our two main methods sample a parent for each vertex in a way that strives to emulate the structure of a DFS traversal while taking into account the given probability distribution. We provide detailed descriptions of our methods in Appendix \ref{sec: Appendix: Randomness in Sampling (DFS)}. 

\subsection{Bellman-Ford}
The Bellman-Ford algorithm computes the minimum-cost paths from a single source vertex $s$ to each other vertex $v$ in the graph. As is the case for DFS, the solutions are represented as a predecessor array $\pi$.

\subsubsection{Randomness in training}%\label{sec: Methodology Randomness in training (BF)}
Similarly to randomising DFS, our implementation of a randomised Bellman-Ford algorithm relies on randomising the order in which paths are considered by shuffling the lists of nodes. %Figure \ref{fig:bf_multpath} illustrates a situation in which multiple valid Bellman-Ford predecessor arrays are possible. However, the final output of the canonical Bellman-Ford algorithm will only be influenced by such a randomisation procedure if there are multiple minimal-weight paths from the source node $s$ to at least one $v \in V \backslash \{s\}$. As the CLRS-30 benchmark samples Erd\H{o}s-Rényi graphs with real-valued edge weights, the probability of two paths $(s,v)$ for any $v \in V \backslash \{s\}$ having the same cost is vanishingly low. To this end, we use a smaller set of possible weights such that for each weight $w_{ij} \in \{1,2,3\}$, which we normalise for training purposes. 
As with DFS, we obtain a distribution of parent nodes by re-running the algorithm 20 times and computing the frequency of each parent for each node (Appendix \ref{sec: Appendix Randomness in training (BF)}).
\begin{figure}[h!]
    \centering
    \includegraphics[width=0.4\linewidth]{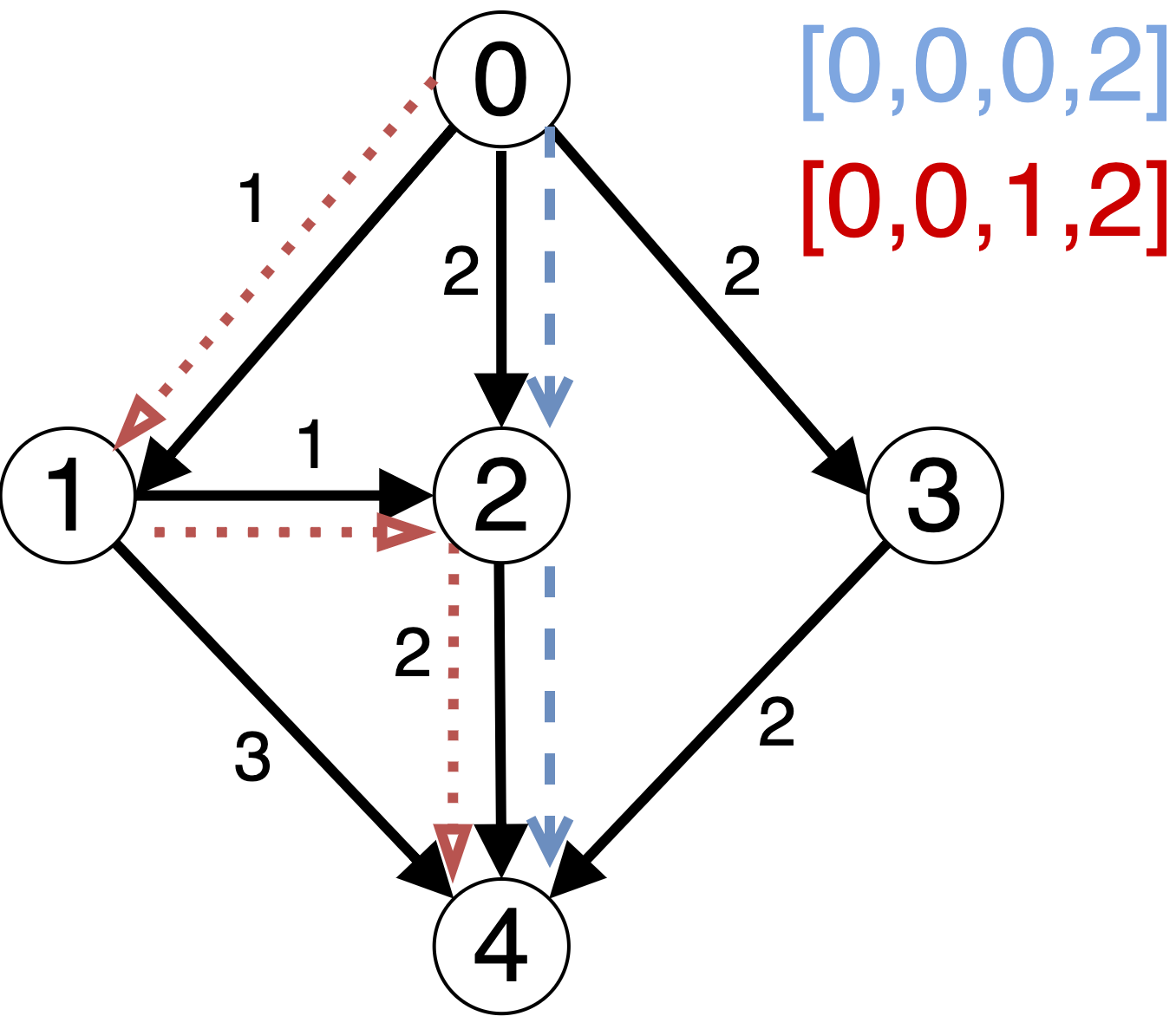}
    \caption{A graph with multiple possible minimum-cost $(source,v)$-paths. As previously, the black arrows represent the edges, with the red and blue edges showing different minimum-cost paths corresponding to their respective predecessor arrays.}
    \label{fig:bf_multpath}
\end{figure}

\subsubsection{Randomness in sampling}%\label{sec: Appendix: Randomness in Sampling (BF)}
We use a sampling algorithm we call \textit{BF Beamsearch} to sample predecessor arrays $\pi$ representing the minimum-cost $(s,v)$-paths for each vertex in the graph. This algorithm closely resembles standard Beamsearch, but it differs as it samples according to the distribution given by the model (\ref{sec: Appendix: Randomness in Sampling (BF)}). %For each $v \in V(G)$, \textit{BF Beamsearch} considers paths from the source to $v$ of length at most $n$ by first sampling candidate predecessors of $v$ and then recursively sampling predecessors of predecessors until the path has length $n$. At each stage, only the 3 lowest-cost paths are kept and used for the next sampling. As soon as the maximum recursion depth is reached, the predecessor that is at the end of the lowest-cost path from the source is chosen as $\pi[v]$. This mimics the operation of the deterministic Bellman-Ford algorithm while respecting randomness and is thus a suitable, albeit involved, way of sampling $\pi$.
To sample predecessor arrays more easily from $\trueparentdist$ and $\modelparentdist$, we also use a greedy randomised algorithm (\ref{sec: Appendix: Randomness in Sampling (BF)}). 

%This algorithm samples a set of parents for each vertex $v$ except the source node (which we assume to be its own parent) according to $\modelparentdist_v$. In the next step, it checks whether any of the sampled parents are plausible (i.e., if there exists an edge $(p, v) \in E(G)$ for the parent $p$). It resamples if there are no plausible parents. If any of the sampled parents are plausible, or if the maximum number of resamplings is reached, it chooses the plausible parent $p$ with the lowest weight $w_{p,v}$ as the predecessor $\pi[v]$. Furthermore, we also use the methods of deterministically choosing the likeliest parent of $v$ according to $\modelparentdist_v$ and randomly uniformly choosing a parent as was done with DFS in order to obtain some baseline results.

\subsection{Permuting Inputs}
We compare our method for generating multiple solutions by predicting solution distributions and stochastically extracting solutions, with another method: permuting the inputs to the standard CLRS benchmark model \cite{clrs-benchmark}. Each input graph may be encoded many times isomorphically by permuting node labels. It is thus natural to wonder whether permuting node labels will lead to a neural network extracting a new solution.

%It is natural to wonder whether
%Fundamentally, a trained NN has a deterministic forward pass: the same inputs will produce the same solution. Fortunately, many algorithms have multiple ways to encode their inputs. For instance, an adjacency matrix has the same structure irrespective of the node labeling (different permutations are isomorphic). }

%For a shortest paths algorithm like Bellman Ford, asking for shortest paths from 0 with G is the same as asking for shortest paths from node 2 with PTGP. For }

% \subsubsection{Permuting Bellman Ford}
% Trained networks predict on feedback.features, consisting of pos, s, A, and adj. We permute everything according to a fixed permutation.

% \subsubsection{Permuting DFS}
% Trained networks predict on pos, A, and adj. We permute everything according to a fixed permutation.

\section{Experiments}
Addressing the lack of evaluation methods for NAR with multiple solutions, we propose metrics to evaluate the validity and diversity of solutions. We regard a solution graph $G'$ to be valid for a randomized algorithm $A$ and a graph $G$ if $G'$ is in the set of possible outputs of $A(G)$. We assess the diversity of solutions by considering the degree to which repeatedly sampling from the same distribution produces different solutions. For a collection of graphs and corresponding solutions, we call the fraction of correct solutions produced by a given method the \textit{accuracy} of the method (referred to as \textit{graph accuracy in \cite{minder2023salsaclrssparsescalablebenchmark}}).
For a single graph, our NN predicts one solution distribution. 
From this distribution, we sample 5 solutions and count the number of distinct and valid solutions. 
For Bellman-Ford, a solution is valid if and only if it gives all shortest paths from the source vertex. For DFS, no straightforward evaluation method is known and we thus present a novel verification algorithm in Appendix \ref{sec:dfs-check}. Figures \ref{fig:parent-dist} and \ref{fig:extract-solutions} illustrate data generation and solution extraction on a graph. We investigate our coverage of the solution space in Appendix \ref{sec: Appendix: validating distributions}. We run our experiments using the CLRS benchmark \cite{clrs-benchmark}, which handles data generation, model training, and model evaluation in a unified way. The CLRS benchmark generates graphs of size 4, 7, 11, 13, and 16 to train our models on. In keeping with established methodology \cite{clrs-benchmark,ibarz2022generalist}, we test our models on graphs within and outside the range of our training sizes. For within-sample testing, we choose $n = |V(G)| = 5$ and $n = 16$ in order to cover the range of our training data graph sizes. For out-of-distribution evaluation, we use the default CLRS benchmark testing size $n = 64$. For each graph size, we test 32 Erd\H{o}s-Rényi random graphs constructed with edge probability 1/2. The models are trained with default CLRS benchmark model parameters (Appendix \ref{sec: model details}) \cite{clrs-benchmark}.
%\begin{figure}
%    \centering
%    \includegraphics{}
%    \caption{CLRS Algorithm Paradigm: datageneration->training->evaluation}
%    \label{fig:enter-label}
%\end{figure}

\subsection{Results for Bellman-Ford}

On small graphs, ($n = 5$), sampling methods extract a single correct minimum-cost path with perfect accuracy from the NN-predicted distributions (Table \ref{tab: unique}, Table \ref{tab: accuracies}). To validate sampling methods, we also extract solutions from the empirical distributions and conclude that our sampling methods are successful across graph sizes. Similar results for empirical and predicted distributions suggest the NN predicts a good distribution for $n = 5$ graphs. The lack of variety in solutions, with sampling methods always extracting the same solution, is expected for small graphs as there is likely only one minimum-cost path from the source to each target. Diversity increases for $n = 16$ (Table \ref{tab: unique}), with the sampling methods producing a correct solution in all cases when sampling from the empirical distributions and in more than $90\%$ of cases when sampling from the model output distributions. In the case of $n = 64$ graphs sampling a predecessor array five times gives a new solution in nearly all cases (Table \ref{tab: unique}). In Appendix \ref{sec: Appendix: validating distributions, BF}, we examine the diversity of the returned solution in comparison to running the randomized version of Bellman-Ford. For larger graphs, Table \ref{tab: accuracies} shows that sampling methods successfully extract correct minimum-cost paths from the empirical distributions (denoted as $\trueparentdist$). However, extraction is often inaccurate from the distributions of solutions predicted by our models (denoted $\modelparentdist$). These results permit us to conclude that our models perform well on small graphs, but have difficulty scaling outside of their training distribution. Comparing our approach with the deterministic algorithmic frameworks of \cite{clrs-benchmark,ibarz2022generalist} is disadvantageous to our method due to its broader scope. Furthermore in \cite{clrs-benchmark,ibarz2022generalist}, node accuracy is used, meaning that a given solution for a given graph and algorithm scores 50\% if half of its parent-successor relationships are the same as in the (single) correct solution for the graph. There is no such optimal solution that we could compare a given solution against in our case and we thus use graph accuracy (see \cite{minder2023salsaclrssparsescalablebenchmark} and Appendix \ref{sec: Appendix Graph Accuracy} for further discussion of the metrics). It is important to note that graph accuracy is much stricter than node accuracy: a method with node accuracy of 92\% would be expected to score only 1\% graph accuracy on a $n=64$ graph. Therefore, it is difficult to compare our results to previous work in NAR.
%worth noting that our BF results are markedly worse than the 99\% micro-F1 score in \cite{clrs-benchmark,ibarz2022generalist}. 
However, our results are encouraging for future work, as we have demonstrated the feasibility of training neural networks to emulate a version of Bellman-Ford that is more powerful than its deterministic variant.

\begin{table}[h!]
\centering
 \caption{Mean proportions of unique and valid solutions when sampling 5 solutions from empirical ($\trueparentdist$) and model output ($\modelparentdist$) distributions for each graph in 5 training runs. Rounded figures.}\label{tab: unique}
	\begin{tabular}{l@{\hskip 3pt}c@{\hskip 3pt}c@{\hskip 3pt}c@{\hskip 3pt}c}
		\multirow{2.5}{*}{} & \multicolumn{4}{c}{Bellman-Ford}  \\
		\cmidrule(lr){2-5} 
		& \textbf{U}niques ($\trueparentdist$) & \textbf{V}alids ($\trueparentdist$) & \textbf{U} ($\modelparentdist$)  & \textbf{V} ($\modelparentdist$) \\
		\midrule
		Greedy (n=5) &$0.20 \pm 0.0$&
$1.00 \pm 0.0$&
$0.20 \pm 0.0$&
$1.00 \pm 0.0$\\
		Beam (n=5) & $0.20 \pm 0.0$&
$1.00 \pm 0.0$&
$0.20 \pm 0.0$&
$1.00 \pm 0.0$ \\
Greedy (n=16) &$0.35 \pm 0.0$&
$1.00 \pm 0.0$&
$0.40 \pm 0.1$&
$0.92 \pm 0.0$ \\
Beam (n=16) &$0.34 \pm 0.0$&
$1.00 \pm 0.0$&
$0.39 \pm 0.0$&
$0.90 \pm 0.0$\\
        Greedy (n=64) & $1.00 \pm 0.0$&
$1.00 \pm 0.0$&
$1.00 \pm 0.0$&
$0.40 \pm 0.2$\\
        Beam (n=64) & $1.00 \pm 0.0$&
$1.00 \pm 0.0$&
$1.00 \pm 0.0$&
$0.54 \pm 0.2$\\\hline \\
  \multirow{2.5}{*}{} & \multicolumn{4}{c}{Depth-First Search}  \\
		\cmidrule(lr){2-5} 
		& \textbf{U}niques ($\trueparentdist$) & \textbf{V}alids ($\trueparentdist$) & \textbf{U} ($\modelparentdist$)  & \textbf{V} ($\modelparentdist$)  \\
		\midrule
  Upwards (n=5) &$0.73 \pm 0.0$&
$0.36 \pm 0.0$&
$0.45 \pm 0.0$&
$0.19 \pm 0.0$\\ 
  AltUpwards (n=5) & $0.60 \pm 0.0$&
$0.80 \pm 0.0$&
$0.66 \pm 0.0$&
$0.77 \pm 0.0$ \\ 
  Upwards (n=16) &$1.00 \pm 0.0$&
$0.10 \pm 0.0$&
$1.00 \pm 0.0$&
$0.01 \pm 0.0$\\ 
  AltUpwards (n=16) &$1.00 \pm 0.0$&
$0.18 \pm 0.0$&
$1.00 \pm 0.0$&
$0.14 \pm 0.0$ \\ 
Upwards (n=64) & $1.00 \pm 0.0$&
$0.02 \pm 0.0$&
$1.00 \pm 0.0$&
$0.01 \pm 0.0$\\
AltUpwards (n=64) & $1.00 \pm 0.0$&
$0.05 \pm 0.0$&
$1.00 \pm 0.0$&
$0.03 \pm 0.0$\\
		\bottomrule
	\end{tabular}
\end{table}

\subsection{Results for Depth-First Search}\label{sec:DFresults}

Our results indicate that even on small graphs, our methods do not achieve a good accuracy (Table \ref{tab: accuracies}). Without modifications, such as hint pointer reversal \cite{hint-reversal}, DFS is a difficult problem for neural algorithmic reasoners \cite{ibarz2022generalist, clrs-benchmark}, and our results point towards the additional difficulty of learning to execute DFS when considering multiple solutions. As in the discussion of our results for Bellman-Ford, we point out that no suitable comparison exists with previous work. In \cite{ibarz2022generalist}, it is indicated the node-level micro-F1 score for DFS does not exceed 60\% for $n=64$ without hint pointer reversal \cite{hint-reversal} (these values being an improvement over the previous state of the art), making graph-level accuracies significantly above 0\% unlikely. Thus, while our reported accuracies are low, they are likely to be in line with previous work in the NAR, with the additional difficulty of retrieving multiple solutions.
Considering the accuracy results and the fact that even on the solution distributions, our sampling methods do not produce correct solutions, we hypothesize that the encoding of multiple solutions into a probability distribution, as presented above, leads to a lack of differentiation between different correct solutions and thus to difficulties retrieving correct solutions. Further evidence for this hypothesis is the observation that there are fewer correct solutions to the single-source shortest path problem on one of the random weighted graphs we sample than there are valid depth-first traversals of that same graph. This is the case because for a given unweighted graph $G$, every path from the root could be included in a depth-first traversal, but only the shortest paths from the root to the vertices can be part of the output of the randomized Bellman-Ford algorithm on a given weighted version of $G$.

\begin{table}[h!]
	\centering
 \caption{Mean accuracies (0 to 1) for all sampling methods, with standard deviations over 5 runs.}\label{tab: accuracies}
	\begin{tabular}{l@{\hskip 3pt}c@{\hskip 3pt}c@{\hskip 3pt}c@{\hskip 3pt}c}
		\multirow{2.5}{*}{} & \multicolumn{4}{c}{Bellman-Ford}  \\
		\cmidrule(lr){2-5}
		& Argmax & Beam & Greedy  & Random \\
		\midrule
		True (n=5) &$1.00\pm0.0$
&$1.00\pm0.0$
&$1.00\pm0.0$
&$0.00\pm0.0$\\
		Model (n=5)    &$1.00\pm0.0$
&$1.00\pm0.0$
&$1.00\pm0.0$
&$0.00\pm0.0$
\\
True (n=16)
&$1.00\pm0.0$
&$1.00\pm0.0$
&$1.00\pm0.0$
&$0.00\pm0.0$\\
Model (n=16)
&$0.98\pm0.0$
&$0.91\pm0.0$
&$0.92\pm0.0$
&$0.00\pm0.0$\\
True (n=64) &$1.00\pm0.0$
&$1.00\pm0.0$
&$1.00\pm0.0$
&$0.00\pm0.0$\\
		Model (n=64) &$0.50\pm0.3$
&$0.56\pm0.2$
&$0.38\pm0.2$
&$0.00\pm0.0$\\ \hline \\
  \multirow{2.5}{*}{} & \multicolumn{4}{c}{Depth-First Search}  \\
		\cmidrule(lr){2-5} 
		& Argmax & AltUpwards & Upwards  & Random \\
		\midrule
        True (n=5) &$0.80\pm0.1$
&$0.90\pm0.1$
&$0.34\pm0.2$
&$0.00\pm0.0$\\
        Model (n=5) &$0.61\pm0.1$
&$0.50\pm0.1$
&$0.16\pm0.0$
&$0.00\pm0.0$\\
True (n=16) &$0.00\pm0.0$
&$0.00\pm0.0$
&$0.08\pm0.0$
&$0.00\pm0.0$\\
        Model (n=16) &$0.00\pm0.0$
&$0.00\pm0.0$
&$0.00\pm0.0$
&$0.00\pm0.0$\\
        True (n=64) &$0.00\pm0.0$
&$0.00\pm0.0$
&$0.01\pm0.0$
&$0.00\pm0.0$
	   \\ Model (n=64)&$0.00\pm0.0$
&$0.00\pm0.0$
&$0.00\pm0.0$
&$0.00\pm0.0$\\
		\bottomrule
	\end{tabular}
\end{table}

%\begin{figure}
%\centering
%\begin{minipage}{.49\textwidth}
    %\includegraphics[width=0.98\linewidth]{l65_project_template/DFS_n5.png}
 %   \captionof{figure}{Accuracies for different methods of sampling DFS trees on graphs of size 5}
%    \label{fig:dfs_n5}
%\end{minipage}%
%\hfill
%\begin{minipage}{.49\textwidth}
    %\includegraphics[width=0.98\linewidth]{l65_project_template/DFS_n64.png}
%    \captionof{figure}{Accuracies for different methods of sampling DFS trees on graphs of size 32}
%    \label{fig:dfs32}
%\end{minipage}
%\end{figure}

\subsection{Results for Permuting}
Permuting node indices either fails to produce new solutions (each time predicting a permutation of the same original solution), or else incurs significant accuracy penalties (Table \ref{tab: perm accuracies}). 
For DFS with ordered restarts, some of the decline is explained because a solution that was correct originally may become incorrect after a permutation. However, for the Bellman-Ford algorithm, the solutions generated by the model remain correct after permuting the vertex indices. Even so, the neural algorithmic reasoner does not appear to be capable of generating any solutions that are correct and non-isomorphic to the single correct solution generated by the model.
%\john{The results for permuting are bad. This makes sense for DFS, since the model is one where order matters for correctness. This is surprising for Bellman Ford. Either way, it suggests NAR multiple solutions is hard, and the other first-impulse method for generating a variety both incur large accuracy penalties.}

\begin{table}[h!]
	\centering
 \caption{Mean accuracies and proportions of distinct solutions (0 to 1) when permuting inputs 5 times, with standard deviations over 5 runs.}\label{tab: perm accuracies}
	\begin{tabular}{l@{\hskip 3pt}c@{\hskip 3pt}c@{\hskip 3pt}c@{\hskip 3pt}c}
		\multirow{2.5}{*}{} & \multicolumn{4}{c}{Accuracy}  \\
		\cmidrule(lr){2-3} \cmidrule(lr){4-5}
		& DFS& Bellman-Ford & \\
		\midrule
(n=5) 
&$0.10\pm0.03$
&$0.30 \pm0.07$
\\
		
(n=16)
&$0.01\pm0.02$
&$0.00\pm0.00$
\\

(n=64)
&$0.00\pm0.00$
&$0.00\pm0.00$
\\ \hline \\
  \multirow{2.5}{*}{} & \multicolumn{4}{c}{Variety}  \\
		\cmidrule(lr){2-3} \cmidrule(lr){4-5}
		& DFS & BF  \\
		\midrule
(n=5)
&$0.90\pm0.09$
&$0.62\pm0.08$
\\

(n=16)
&$1.00\pm0.00$
&$0.99\pm0.01$
\\

(n=64)
&$1.00\pm0.00$
&$1.00\pm0.00$
\\
		\bottomrule
	\end{tabular}
\end{table}
\section{Conclusion}
We outline a two-part approach to NAR with multiple solutions: we train models to predict distributions of solutions and stochastically extract solutions from those distributions. In the case of using a NAR to find multiple single-source shortest path solutions in a weighted graph we demonstrate that this is feasible, particularly on graphs with sizes that lie within the range of sizes we train our models on. For DFS, we see that this task is significantly more challenging, as indicated by prior NAR research. We attribute this result to a number of factors. First, previous work indicates that emulating DFS with NAR is a difficult task. Second, training a model to fit a distribution of solutions does not directly optimize for correct solutions. Solutions must still be extracted from the distribution via stochastic methods, to which the loss metric does not extend (Figure \ref{fig:num_reruns}). Improving methods for extracting solutions remains an important algorithmic task. Still, with the promise of sampling multiple solutions from a distribution demonstrated, we believe our modifications of the CLRS benchmark act as proof-of-concept for further work. Our method would be applicable to other CLRS benchmark algorithms such as Minimum Spanning Tree, Bipartite Matching, and Task Scheduling. Having multiple solutions adds essential flexibility to applications of NAR such as cyberphysical systems and might aid in explaining NAR models.

\bibliographystyle{ACM-Reference-Format}
\bibliography{log_2022}
\appendix

%\section*{Appendix}
\newpage
\section{Methodology}\label{sec: Appendix: Methodology}
We discuss our methods for randomising the deterministic (canonical) versions of Depth-First Search and the Bellman-Ford algorithm in order to use the resulting distributions over solutions as inputs to the CLRS baseline. We also present our methods of extracting candidate solutions from model outputs and give approaches towards verifying them as correct solutions of the respective deterministic algorithms our sampling methods are aiming to emulate.

\begin{figure}[H]
    \includegraphics[width=0.8\linewidth]{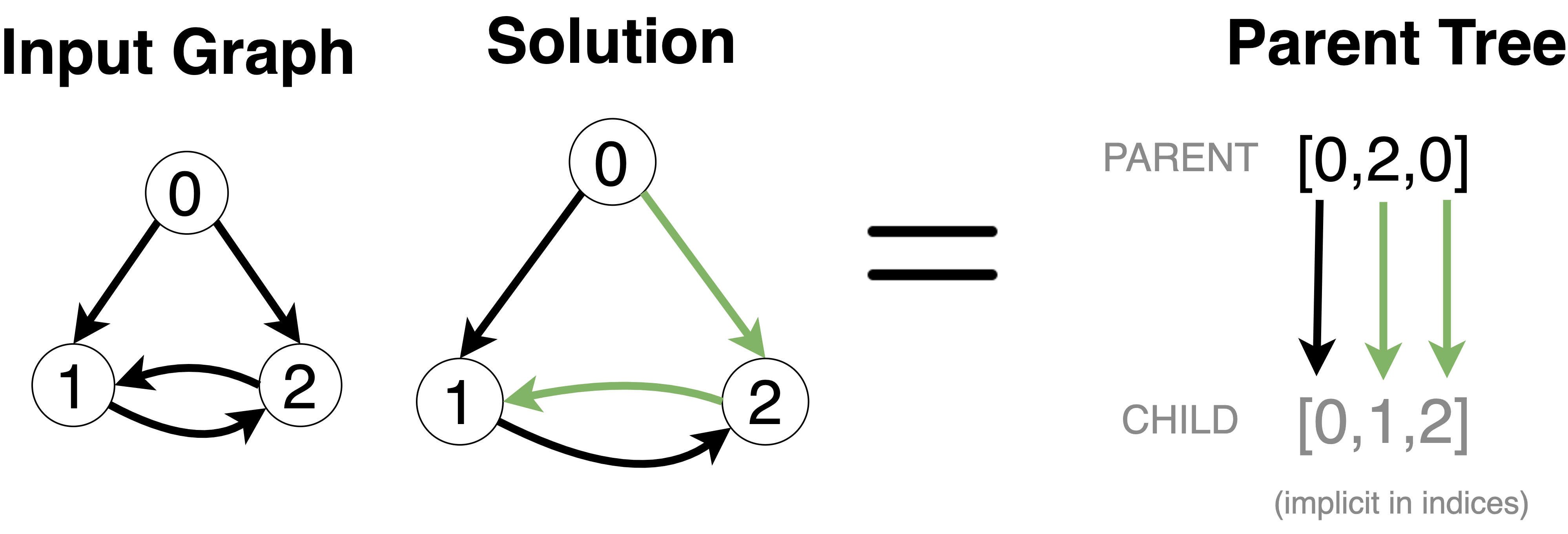}
    \caption{Parent Tree encoding of a single DFS solution. The start node, 0, is its own parent, represented by the value 0 at index 0. The value 2 at index 1 indicates that vertex 2 is the parent of vertex 1. Similarly, the value 0 at index 2 indicates that vertex 0 is the parent of vertex 2.} %Henceforth we draw only the parent list: the children are implicit in the indices.
    \label{fig:enter-label_app}
\end{figure}

%Find the thing you can shuffle. Shuffle it. 
%Introduce randomness in training, Model generates probabilities that we hope encode multiple solution,Sample with randomness from model probabilities to generate a solution. Sampling multiple times will give multiple solutions.
%Check validity. Safety is good.
\begin{figure}[H]
    \centering
    \includegraphics[width=0.8\linewidth]{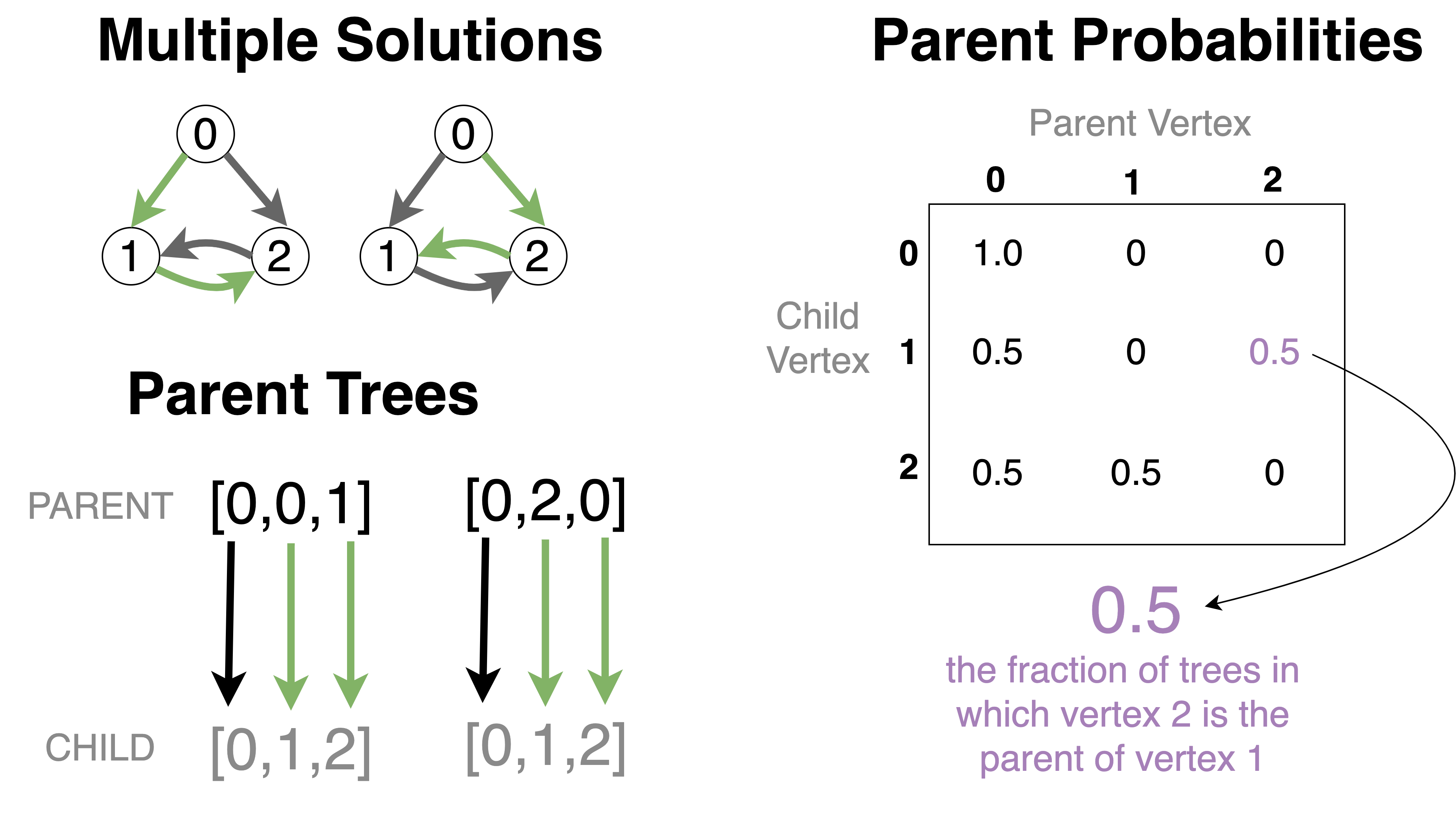}
    \caption{Parent Probability Distribution encoding of multiple DFS solutions. The value 0.5 in row $1$ column $2$ indicates that vertex 2 is a parent of vertex 1 in 50\% of correct parent trees. We call the parent probabilities obtained by repeatedly running an algorithm the \textit{true parent distribution} $\trueparentdist$. The model outputs, which we aim to be as close as possible to $\trueparentdist$, are denoted as $\modelparentdist$.} 
    %\label{fig:parent-dist}
\end{figure}
%\footnote{The distribution of parent probabilities, equally called the solution distribution, can also be thought-of as a weighted digraph, where an edge $(s,t)$ with weight 0.5 indicates a 50\% probability that vertex $t$ is the parent of vertex $s$.} %\footnote{Note $\trueparentdist$ is an approximation.}

\begin{figure}[H]
    \includegraphics[width=\linewidth]{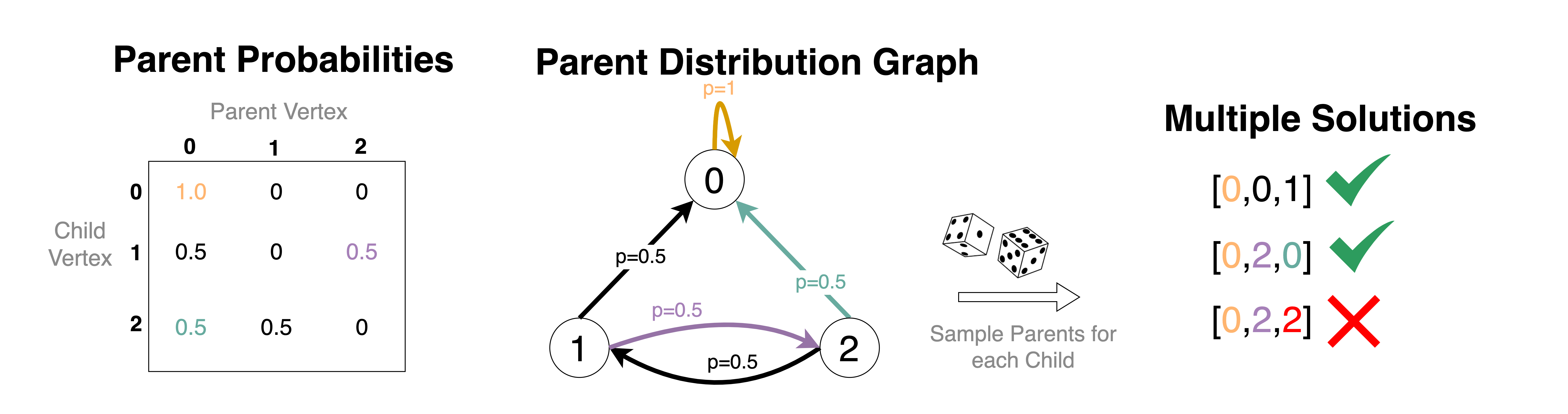}
    \caption{DFS tree extraction from a parent probability matrix:  $\modelparentdist$ or $\trueparentdist$. The parent tree [0,2,2] is incorrect because there is no solution in which 2 is its own parent. [0,0,0] is incorrect, despite there being solutions in which 0 is the parent of each vertex, because there is no solution in which 0 is the parent of every vertex: if 0 is the parent of 1, 1 must be the parent of 2.}
    
    %\zeno{would cut this: Extracting Multiple Parent Trees from a Distribution of Parent Probabilities. The matrix of parent probabilities can be seen as an adjacency matrix for a weighted digraph where edges point from child to parent.} The process from a parent probability matrix $\trueparentdist$ or $\modelparentdist$  to a DFS tree. \zeno{would cut this: Walking with randomness across the graph from child to parent produces a solution, in this case a parent tree.} The correctness of a parent tree depends on the original input graph and the specific algorithm \zeno{would cut: (a dfs parent tree has different conditions for correctness than a single-source shortest paths parent tree)}. In the case of DFS, [0,2,2] is invalid since there is no solution in the solution distribution with an edge from 2 to itself, but [0,0,0] would also be incorrect, even though all nodes have edges to 0, as DFS forbids backtracking to 0 until paths outgoing from 1 have been explored. This provides an example of why validation of our sampled trees is critical.}
    %\label{fig:extract-solutions}
\end{figure}

\subsection{DFS}
We consider a generalization of the deterministic DFS setting implemented in the CLRS benchmark \cite{clrs-benchmark}. In our setting, edge-exploration is random: from a given vertex, any of the outgoing edges may be explored first. Still, we require that once DFS backtracks at the root of a connected component, it restarts its search at the next vertex in a linear ordering of the vertices. We think the generalization is natural, modelling a setting where one conducts DFS according to a chosen sequence of starting points, without assumptions about which edges to pursue first.

%Broadening to a DFS setting where any outgoing edge might be explored first allows us to consider multiple solutions correct. We now say this in figure captions, right? Figure \ref{fig:dfs_multpath gives a simple example of a situation where multiple correct DFS-trees exist when randomising the order of exploration while keeping the start node fixed.} Keeping the property that connected components will be explored according to a linear order helps us determine the correctness of a proposed solution along necessary conditions we introduce. We also think it has the advantage of naturality, modelling a setting where one conducts DFS along connected components according to an inductive bias of starting points.

\begin{figure}[H]
    \includegraphics[width=0.4\linewidth]{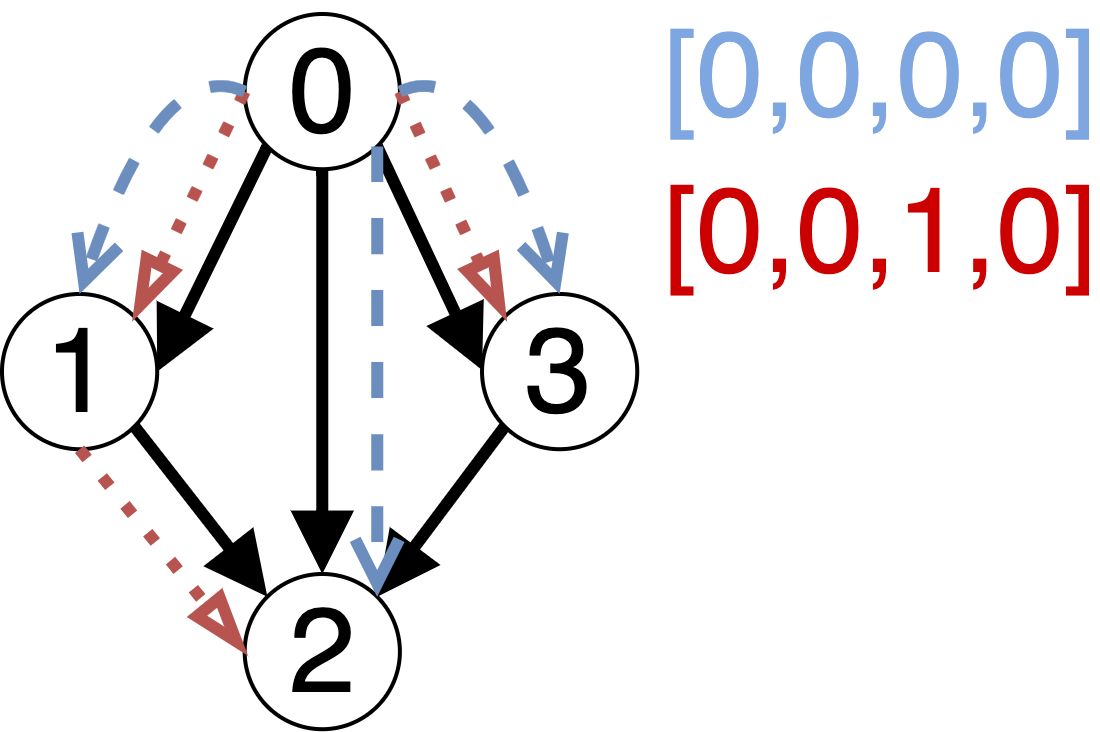}
    \caption{A graph with multiple correct DFS solutions. Graph edges are drawn in black. The red tree would be the result if vertices were explored in alphabetical order, while the blue tree is one of multiple other posible valid DFS trees. Note also how the DFS-tree is encoded as a predecessor array (with the entry at position $i$ is the predecessor of the $i^{th}$ vertex in alphabetical order, which we henceforth call $\pi$.}
    \label{fig:dfs_multpath_app}
\end{figure}

\subsubsection{Randomness in training}\label{sec: Appendix Randomness in training (DFS)}
Algorithm \ref{alg:dfsOrdered} gives detailed description of the adapted randomized DFS algorithm used in this paper, with the line that randomizes edge-exploration highlighted in red. After running Algorithm \ref{alg:dfsOrdered} 20 times, we compute the fraction of solutions in which node $s$ is the parent of node $t$, which we say is the parent probability, as shown in Figure \ref{fig:parent-dist}.
\begin{algorithm}
\caption{Randomised Depth-First Search}\label{alg:dfsOrdered}
\begin{algorithmic}[1]
    \State Input: Adjacency Matrix $A$, vertices $V$ in linear order
    \State Output: Parent Tree $\pi$, encoded as list: $\pi[i] := $ parent of $i$
    \State Initialize color $\gets [0]*|V|$ 
    \State Initialize startNode $\gets V[0]$
    \State rootOrdering $\gets$ $V$ \Comment{ordered restarts}
    \State \textcolor{red}{tiebreakOrder $\gets$ shuffle$(V \setminus V[0])$} \Comment{random edge-exploration}
    \For{\upshape{root in rootOrder}}
        \If{color[root] == 0} \Comment{vertex not yet explored}
            \State color[root] $\gets 1$
            \State $\pi$[root] $\gets$ root
            %% explore then backtrack.
            \State current $\gets$ root \Comment{pointer to node being explored}
            \State $s_{last} \gets$ root  \Comment{pointer to node most-recently found}
            \State $s_{prev}[s_{last}] \gets$ root \Comment{pointer to node before the node most-recently found}
            
            \While{True}
                
                \For{\textcolor{red}{\upshape{potentialChild in tiebreakOrder}}}
                    \If{A[root][potentialChild] != 0}
                        \If{color[potentialChild == 0]}
                            \State color[potentialChild] $\gets$ 1 \;
                            \State $\pi$[potentialChild] $\gets$ current\;
                            \State $s_{last} \gets potentialChild$ \;
                            \State break\; \Comment{child found, exit for-loop, continue while}
                        \EndIf
                    \EndIf
                \EndFor
                
                \If{current == $s_{last}$} \Comment{No outgoing edge found}
                    \If{$s_{prev}[s_{last}] == s_{last}$} \Comment{Done: backtracked source and no new edges}
                        \State break \Comment{Exit while loop, try further with next root in rootOrder}
                    \EndIf
                    \State $s_{last} \gets s_{prev}[current]$ \Comment{Prepare to Backtrack}
                \EndIf
                \State current $\gets s_{last}$ \Comment{go-deeper if new node found, else backtrack}
            \EndWhile
        \EndIf
    \EndFor
\end{algorithmic}
\end{algorithm}

\subsubsection{Randomness in sampling}\label{sec: Appendix: Randomness in Sampling (DFS)}
%To address the aim of the project, it is critical to solve the problem of obtaining valid DFS solutions from the distribution $\modelparentdist$ of parents. This section concerns itself with sampling methods, while the next section describes some necessary conditions against which sampled solutions are checked. The first method we employ is not strictly sampling, as we simply assume the parent of each node $x$ to be the node $u$ such that:

We compare 4 methods for extracting valid DFS solutions from a parent distribution, $\modelparentdist$. First, argmax sampling assumes the parent of each node $x$ to be the node $u$ such that
\begin{equation}
    u = \underset{v \in V(G)}{\text{max}} \modelparentdist_{v,u},
\end{equation}
where $\modelparentdist_{v,u}$ is the probability of an edge $(v,u)$ according to $\modelparentdist$. Argmax sampling is deterministic and can return only one solution. We include it in our discussion as a good baseline which corresponds to the internal mechanism of the CLRS-30 benchmark \cite{clrs-benchmark}. To extract multiple solutions for DFS, we introduce stochasticity according to \textit{Upwards Sampling}, which is described in Algorithm \ref{alg:upwards}.

%Given that the likeliest predecessor node is deterministic given $\modelparentdist$, this method hi
%Another baseline we employ is random uniform sampling of predecessor arrays.we don't display this result %None of the above methods however pays credence to operation of DFS. To this end, we introduce \textit{Upwards Sampling}, which is described in Algorithm \ref{alg:upwards}. This method is implemented to handle entire test set batches in our code, but we restrict our attention to a single example for clarity in the pseudocode.
%Note that we sample $\pi$ from $\trueparentdist$ identically for comparison.
%For later use, we also define the notation $\trueparentdist_v$ to denote the probability distribution over the possible parents of a vertex $v$. never used!
\begin{algorithm}
\caption{Upwards Sampling}\label{alg:upwards}
\begin{algorithmic}[1]
    \State Input: $\modelparentdist$
    \State Output: A sampled predecessor array $\pi$
    \State Initialize $\pi$
    \State leaves $\gets sort(\sum_{v \in V} \modelparentdist_{u,v} \text{ for all } u \in V)$ \label{lin:leafiness} \Comment{Sort vertices by their probability of being leafs}
    \While{$leaves \neq \emptyset$} \label{lin:outerloop}
    \State leaf $\gets leaves[0]$ \Comment{Take the vertex most likely to be a leaf} 
    \State $\pi[leaf] \gets SamplePredecessor(\modelparentdist,leaf)$ \label{lin:sampleparent}\Comment{Sample $\pi[leaf]$ according to $\modelparentdist_{leaf}$}
    \State leaves $\gets leaves.remove(leaf)$ \Comment{Prevent leaf from being considered as a child again}
    \For{$v \in V$} \State $\modelparentdist_{leaf,v} \gets 0$ \Comment{Prevent leaf from being sampled as a parent} 
    \EndFor
    \State leaf $\gets \pi[leaf]$ \Comment{Make predecessor the next leaf}
    \While{$\pi[leaf]$ is undefined}\label{lin:innerloop} \Comment{Sample the predecessor's predecessor}
    \State $\pi[leaf] \gets SamplePredecessor(\modelparentdist,leaf)$ 
    \State leaves $\gets leaves.remove(leaf)$ 
    \For{$v \in V$} \State $\modelparentdist_{leaf,v} \gets 0$ 
    \EndFor
    \State leaf $\gets \pi[leaf]$
    
    \EndWhile
    \EndWhile
\end{algorithmic}
\end{algorithm}

%We now turn to a more detailed discussion of Upwards Sampling. 
In Upwards Sampling (Algorithm \ref{alg:upwards}), we first determine how likely each node is to be a parent node of any other node. We do so by summing the probabilities of each vertex $u \in V$ to be the parent of each vertex $v \in V$ and then sorting this list in ascending order, as done in Line \ref{lin:leafiness}. In each iteration of the outer loop (Line \ref{lin:outerloop}), we begin by taking the vertex with the lowest probability of being a parent and sampling its parent according to $\modelparentdist_{leaf}$ (Line \ref{lin:sampleparent}). After having done so, we remove the leaf from future consideration as a leaf and as a parent (as a leaf by definition will not be the parent of another node). Next, we sample the parent of $\pi[leaf]$ until we encounter a node that already has a parent (Line \ref{lin:outerloop}). This ensures we do not overwrite previously sampled parents. We have chosen this sampling method as it works similarly to DFS while respecting the inherent stochasticity resulting from returning $\modelparentdist$ rather than one unambigous solution, as has been done in \cite{clrs-benchmark} and \cite{ibarz2022generalist}. During testing, we encountered the situation that the removal of leaves from consideration as parents leads to no potential parents being available for a node due to having been removed. To address this problem, we give a node a random parent in this case. We further implement a method without this removal, which we call \textit{AltUpwards Sampling}. While less faithful to graph properties, it addresses the limitations caused by the difficulty of computing the likelihood of being a leaf for each node while operating closer to DFS. It can be seen in Section \ref{sec:DFresults} that this method outperforms Algorithm \ref{alg:upwards}.

\subsubsection{Checking Validity} \label{sec:dfs-check}

We verify whether a given forest $F$ is the result of some depth-first traversal of the graph with ordered restarts (RDFSO). By RDFSO, we mean DFS with randomised tiebreaking and ordered restarts. Randomised tiebreaking means that given two successors $v_1,v_2$ of a node $v \in G$, the algorithm chooses randomly from which node to continue the depth-first traversal. Ordered restarts means that when all descendants of a given root have been visited by the search, the algorithm continues DFS from the lowest-indexed unvisited node. Since we are not aware of a solution to this problem in the literature, we define it as the randomised DFS recognition problem (RDFSR) and present an algorithm to solve it.

\begin{theorem}
    Algorithm \ref{Alg:Henry} accepts if and only if a forest $F$ is the output of some run of randomised DFS with ordered restarts (RDFSO) on $G$.
    
    \begin{idea}
        The algorithm attempts to build a valid DFS traversal of $G$, using $F$ to inform the traversal order. We derive the traversal order by comparing the descendants of each node in the graph $G$ and forest $F$. The key fact is that any depth-first traversal from node $v$ visits all unvisited nodes reachable from $v$ before backtracking to $v$. As a result, \textbf{all unvisited nodes reachable from $v$ in graph $G$ are still reachable from $v$ in DFS-forest $F$}. We use this to rebuild a DFS traversal for $F$ one vertex at a time, visiting next whichever vertex has the same unvisited descendants in $G$ that it does in $F$. 
        
        %Conversely, if the descendants of a node $v$ are different in $G$ and $F$, we must wait for another node before $v$.
        
        %Therefore, for a node $s$ with two children $v,w$ with $D_G(v) \cap D_G(w) \neq \emptyset$, we can ascertain whether the traversal continues to $v$ or $w$ by checking if $D_G(v) = D_F(v)$ or $D_G(w) = D_F(w)$ (if neither is the case, we know that $F$ cannot be a valid result of RDFSO$(G)$). Note that if $D_G(v) \cap D_G(w) = \emptyset$, the output of $F$ is invariant to the order in which the traversal continues from $w$. On a high level, our algorithm establishes the traversal order that must have produced the given $F$ by recursively checking descendant sets and rejecting if the descendant sets are not consistent with any run of RDFSO($G$).
    \end{idea}

    \end{theorem}

        We present Algorithm \ref{Alg:Henry}, which solves DFS verification by rebuilding $F$ by traversing $G$. 
        Algorithm \ref{Alg:Henry} rebuilds $F$ one vertex at a time, removing each vertex after considering it. 
        Algorithm \ref{Alg:Henry} discovers the next vertex in traversal by calling Algorithm \ref{Alg: CCV}.
        
        %We present Algorithm \ref{Alg:Henry}, which initializes the array $visited$ through which the algorithm keeps track of which vertices have been visited during the rebuilding of $F$. Algorithm \ref{Alg:Henry} then uses Algorithm \ref{Alg: CCV} as a subroutine to recursively establish whether there is a tiebreak order under which $F$ could have arisen as the output of running RDFSO on $G$. %Note that once a vertex is marked as visited, we remove it from $G$ to update reachability — just as without backtracking to $v$. 

        Algorithm \ref{Alg: CCV}, the subroutine called by our main algorithm, finds the next vertex after $v$ in the traversal order by looking for the adjacent vertex to $v$ in $F$ that discovers all its descendants in $G$. 
        If our most recently visited node $m$ has two children $v_1, v_2$ in $F$, Algorithm \ref{Alg: CCV} determines whether traversal can continue first to $v_1$ or $v_2$ and still produce $F$. 
        %The subroutine algorithm \ref{Alg: CCV} rejects if the input vertex $v$ has different descendants in $F$ and $G$, meaning $v$ is not a valid next vertex from which a DFS traversal that produced $F$ from $G$ could have begun and accepts otherwise. Since we assume ordered restarts, we begin the check by calling Algorithm \ref{Alg: CCV} on vertex 0. The state of the traversal is given by the markings stored in $visited$. If Algorithm \ref{Alg: CCV} does not reject after being called successively on all vertices, Algorithm \ref{Alg:Henry} accepts.

        Line \ref{Alg: DVC, secondcond} checks whether the vertex $v$ is indeed a valid next vertex for the traversal. We write $D_G(v)$ to mean the descendants of node $v$ in graph $G$. Line \ref{Alg: DVC, secondcond} checks whether $D_G(v) = D_F(v)$.
        
        %In Line \ref{Alg: DVC, firstcond}, Algorithm \ref{Alg: CCV} handles the base case of the recursion and Line \ref{Alg: DVC, secondcond} checks whether the vertex $v$ is indeed a valid next vertex for the traversal (by checking whether $D_G(v) = D_F(v)$ when taking into account the $visited$ array, as discussed above. 
        
        The loop in Lines \ref{lin: establish order start}-\ref{lin: establish order end} of Algorithm \ref{Alg: CCV} finds the first vertex $v_1$ adjacent to $v$ for which $D_G(v_1) = D_F(v_1)$ and recursively repeats this procedure to retrace the run of RDFSO until it reaches the maximum depth in $G$, removing the vertices discovered during the recursive calls in $G$ and $F$, thus rendering them unreachable in $G$ from other vertices. It then continues the loop in Lines \ref{lin: establish order start}-\ref{lin: establish order end} with the remaining descendants of $v$. 

        Say there is a vertex $v_2$ adjacent to $v$ in $F$ from which the depth-first traversal has continued after having terminated its traversal from $v_1$. After the vertices discovered by $v_1$ have been removed from $G$, the sets of vertices reachable from $v_2$ in $F$ and $G$ must now be the same.
        %It thus finds the vertex $v_2$ adjacent to $v$ in $F$ from which the depth-first traversal has continued after having terminated its traversal from $v_1$: after the vertices discovered by $v_1$ have been removed from $G$, the sets of vertices reachable from $v_2$ in $F$ and $G$ must now be the same. 
        This is the case because if $D_G(v_1) \cap D_G(v_2) \neq \emptyset$ before the traversal, then all vertices in $D_G(v_1) \cap D_G(v_2)$ have been removed during the call from $v_1$ (or during an earlier call) and are now unreachable from $v_2$ in $G$. Our algorithm now picks $v_2$ as the next vertex from which to continue the traversal. Therefore, if Algorithm \ref{Alg: CCV} has not rejected previously, then the set of vertices reachable from $v_2$ in $F$ and $G$ at a given point could have only been discovered in a traversal from $v_2$. We thus continue the recursive traversal from $v_2$, et cetera.  
        
        If the algorithm encounters a situation where none of the remaining descendants of a given vertex can reach the same set of reachable vertices in $F$ and $G$, the algorithm rejects. This is equivalent to the idea that there is no vertex the traversal can continue towards and still result in $F$. Therefore, the algorithm rejects if there is ever no vertex that discovers all its descendants in $G$ during the run of RDFSO and accepts only if such a vertex exists at every depth of the recursion.

        % PROOF: suppose F was indeed produced by RDFS on G. Then, there was some DFS node-order traversal v_1,v_2,...,v_n. Note at the discovery of v_k, all unvisited nodes reachable from v_k will be discovered before backtracking past v_k (this is exactly what it means to be depth-first). So, all unvisited nodes reachable from v_k will be descendants of v_k in F. 
        
        %Consider now the execution of Algorithm \ref{Alg: Henry} on (F,G). First, Algorithm \ref{Alg: Henry} will check v_1, ascertain that D_F(v_1) = D_G(v_1) and mark v_1 as visited. It will iterate through D_F(v_1) until it finds v_b with D_F(v_b) = D_G(v_b)... then it will recursively call \ref{Alg: CCV} marking the descendants of v_b
        % it wont necessarily traverse v_1,v_2,...,v_n but it will traverse in a way that replicates $F$

        % suppose in the other direction that Algorithm \ref{Alg: Henry} returns True. Then, all nodes were marked visited
        %there was some traversal order that produced F. The traversal order was depth-first (bcuz of DVC recursive calls and the loop through all unvisited descendants). The traversal order had ordered restarts (bcuz of the outer loop of \ref{Alg: Henry}. So, Algorithm \ref{Alg: Henry} found some RDFS traversal order on $G$ that produced $F$, which means F is producible by RDFS on G.

\begin{algorithm}
\caption{RDFSO Verifier} \label{Alg:Henry}
\begin{algorithmic}[1]
\State \textbf{Input: } $G,F$
\For{each vertex $i$ in $G$}
    \If{not DVC($G$, $F$, $i$)}
        \State \Return False
    \EndIf
\EndFor
\State \Return True
\end{algorithmic}
\end{algorithm}

\begin{algorithm}
\caption{Descendant Validity Check (DVC)} \label{Alg: CCV}
\begin{algorithmic}[1]
\State \textbf{Input: } $G, F, \text{vertex }v$

\If{$D_G(v) \neq D_F(v)$} \label{Alg: DVC, secondcond}
    \State \Return False
\EndIf
\State $descendants \gets$ descendants of $v$ in $F$ \label{lin: compute unvisited F-descendants}
\State Delete $v$ from $G$ and $F$
%\While{$descendants$ is not empty}
\For{each $descendant$ in $descendants$} \label{lin: establish order start}
    \State Compute $pd[descendant]$ \Comment{compute the possible descendants in $G$}
    \State Compute $ad[descendant]$ \Comment{compute the actual descendants in $F$}
    \If{$pd[descendant] = ad[descendant]$} \Comment{next vertex in traversal found}
        \If{not DVC($G$, $F$, $descendant$)}
            \State \Return False
        \EndIf
        \State restart line \ref{lin: establish order start}, with updated $descendants$ \Comment{vertices may have been deleted}
    \Else
        \State Check next descendant
    \EndIf
\EndFor \label{lin: establish order end}
\State \Return True if $descendants$ is empty. Else return False.
\end{algorithmic}
\end{algorithm}

\subsection{Bellman-Ford}
The Bellman-Ford algorithm computes the minimum-cost paths from a single source vertex $s$ to each other vertex $v$ in the graph. As is the case for DFS, the solutions are represented as a predecessor array $\pi$.

\subsubsection{Randomness in training}\label{sec: Appendix Randomness in training (BF)}
\begin{figure}[ht]
    \centering
    \includegraphics[width=0.4\linewidth]{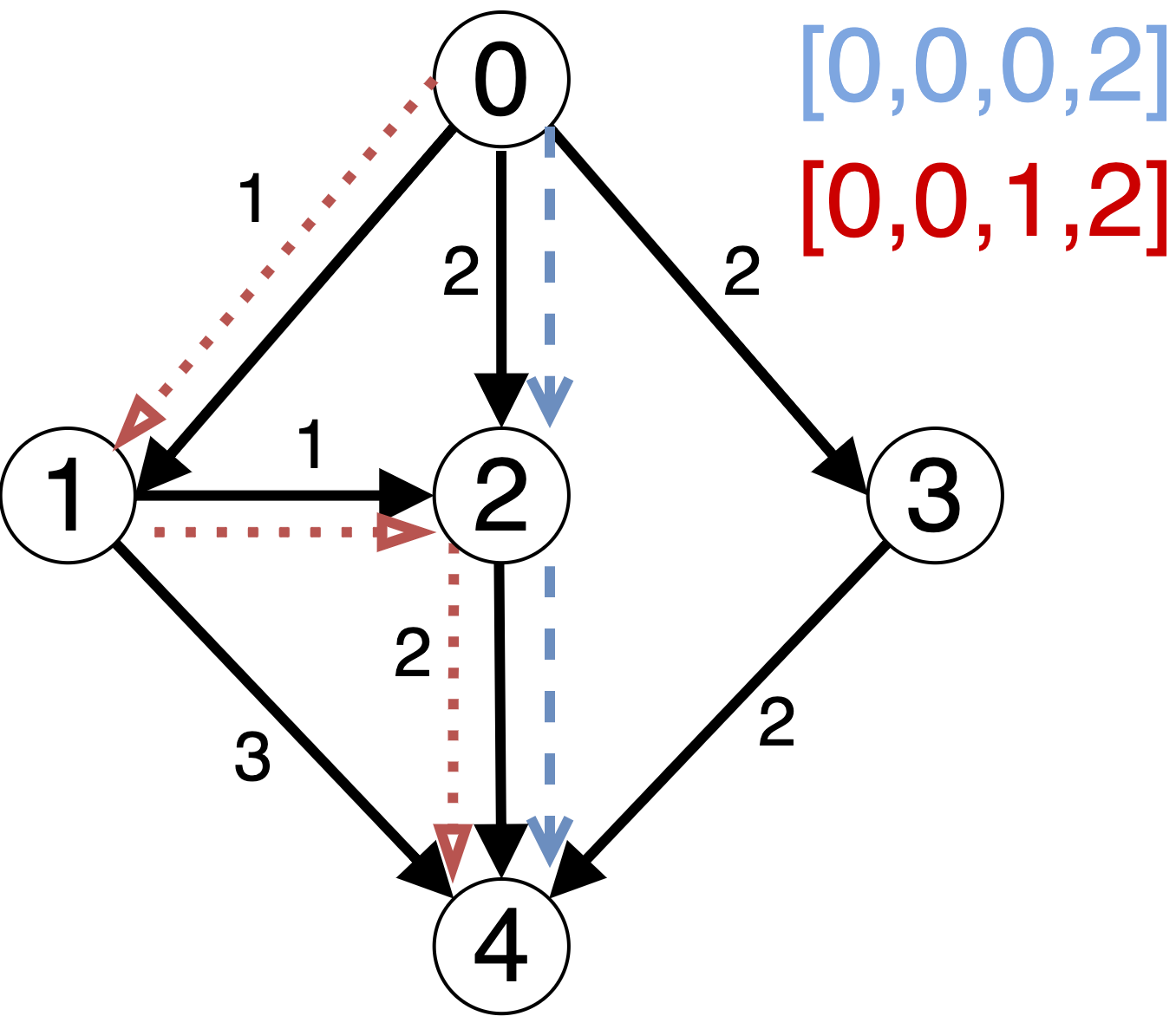}
    \caption{A graph with multiple possible minimum-cost $(source,v)$-paths. As previously, the black arrows represent the edges, with the red and blue edges showing different minimum-cost paths corresponding to their respective predecessor arrays.}
    \label{fig:bf_multpath_app}
\end{figure}

Similarly to randomising DFS, our implementation of a randomised Bellman-Ford algorithm relies on randomising the order in which paths are considered by shuffling the lists of nodes. Figure \ref{fig:bf_multpath_app} illustrates a situation in which multiple valid Bellman-Ford predecessor arrays are possible. However, the final output of the canonical Bellman-Ford algorithm will only be influenced by such a randomisation procedure if there are multiple minimal-weight paths from the source node $s$ to at least one $v \in V \backslash \{s\}$. As the CLRS-30 benchmark samples Erd\H{o}s-Rényi graphs with real-valued edge weights, the probability of two paths $(s,v)$ for any $v \in V \backslash \{s\}$ having the same cost is vanishingly low. To this end, we use a smaller set of possible weights such that for each weight $w_{ij} \in \{1,2,3\}$, which we normalise for training purposes. As for DFS, we obtain a distribution of parent nodes by re-running the algorithm 20 times and computing the frequency of each parent for each node.

\subsubsection{Randomness in sampling}\label{sec: Appendix: Randomness in Sampling (BF)}
For each $v \in V(G)$, \textit{BF Beamsearch} considers paths from the source to $v$ of length at most $n$ by first sampling candidate predecessors of $v$ and then recursively sampling predecessors of predecessors until the path has length $n$. At each stage, only the 3 lowest-cost paths are kept and used for the next sampling. As soon as the maximum recursion depth is reached, the predecessor that is at the end of the lowest-cost path from the source is chosen as $\pi[v]$. This mimics the operation of the deterministic Bellman-Ford algorithm while respecting randomness and is thus a suitable, albeit involved, way of sampling $\pi$.
To sample predecessor arrays more easily from $\trueparentdist$ and $\modelparentdist$, we use a greedy randomised algorithm. This algorithm samples a set of parents for each vertex $v$ except the source node (which we assume to be its own parent) according to $\modelparentdist_v$. In the next step, it checks whether any of the sampled parents are plausible (i.e., if there exists an edge $(p, v) \in E(G)$ for the parent $p$). It resamples if there are no plausible parents. If any of the sampled parents are plausible, or if the maximum number of resamplings is reached, it chooses the plausible parent $p$ with the lowest weight $w_{p,v}$ as the predecessor $\pi[v]$. Furthermore, we also use the methods of deterministically choosing the likeliest parent of $v$ according to $\modelparentdist_v$ and randomly uniformly choosing a parent as was done with DFS in order to obtain some baseline results.

\subsubsection{Checking Validity}
\label{sec:bf-check}
In contrast to DFS, the task of deciding whether a predecessor array sampled from a model output distribution $\modelparentdist$ is a valid output of running the Bellman-Ford algorithm is straightforward. To this end, we implement a simple algorithm called \textit{check valid BF path}, which is given in Algorithm \ref{alg:checkbf}.

\begin{algorithm}
\caption{check valid BF path}\label{alg:checkbf}
\begin{algorithmic}[1]
    \State Input: Adjacency Matrix $A$, source vertex $s$, sampled predecessor array $\pi$
    \State Output: Decide whether $\pi$ is a correct Bellman-Ford solution
    \State $G' \gets buildGraph(\pi)$ \Comment{Construct a graph given by the paths in $\pi$}
    \If{$E(G') \not\subseteq E(G)$} \Comment{Check whether $G'$ contains any nonexistent edges}
    \State Return False
    \Else
        \State $true\_costs \gets BellmanFord(A,s)$ \Comment{Compute the costs using the Bellman-Ford algorithm}
        \State $model\_costs = PathCosts(\pi)$\Comment{Compute costs of paths in $\pi$}
        \State Return $model\_costs == true\_costs$
    \EndIf
\end{algorithmic}
\end{algorithm}

Algorithm \ref{alg:checkbf} simply constructs a graph from the vertices and edges that are implicitly given in $\pi$ and checks whether all edges correspond to edges in the graph $G$ and whether the costs of the paths in $\pi$ are the same of the costs when using the deterministic Bellman-Ford algorithm on $(G,s)$.
%\newpage
\section{Using different numbers of algorithm re-runs}\label{sec: test re-run numbers}
\begin{figure}[H]
  \centering
  \begin{subfigure}[]{0.45\textwidth}
    \centering
    \includegraphics[width=\textwidth]{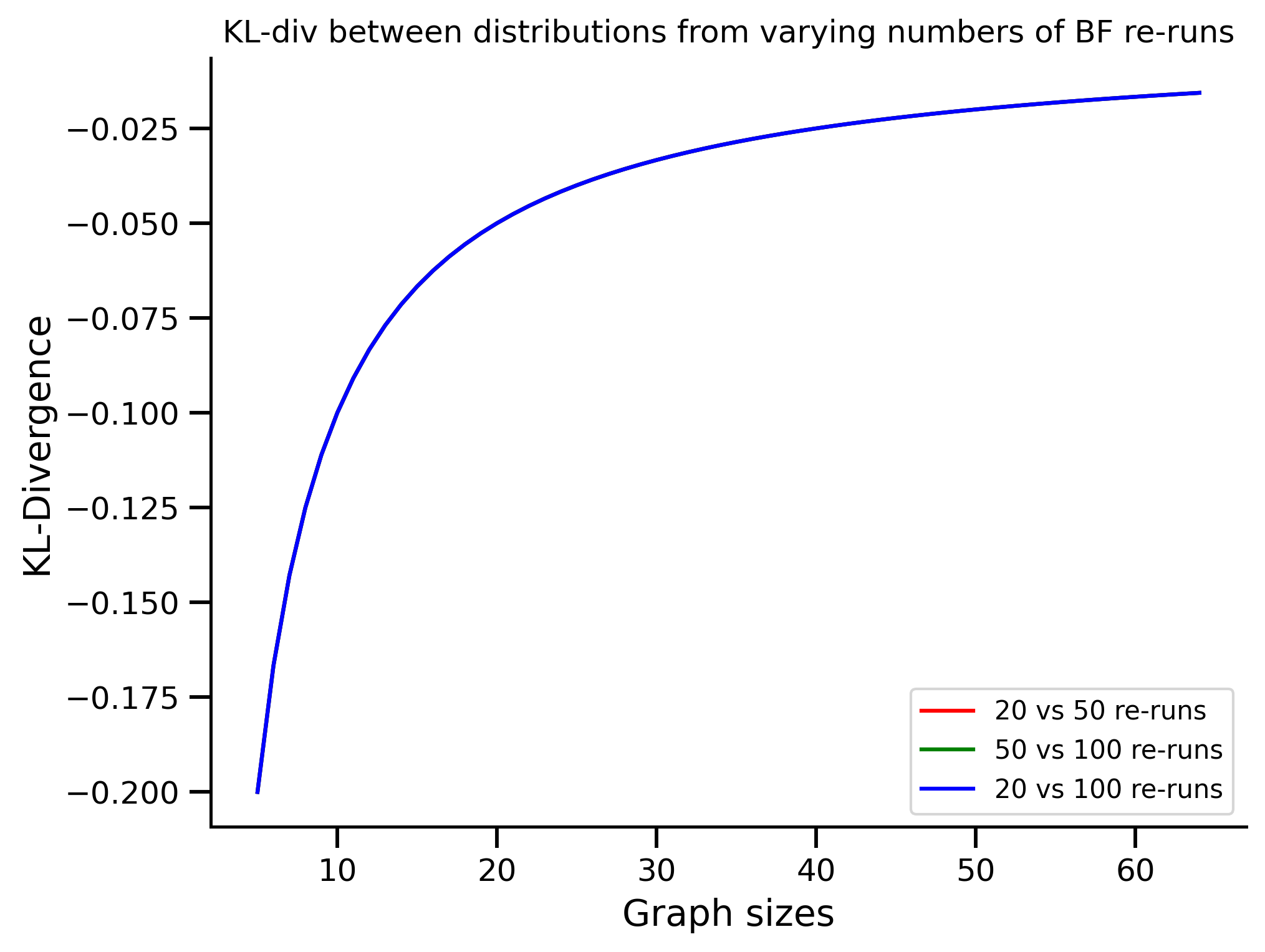}
    \caption{Bellman-Ford}
    \label{fig:bf_sampling_n5}
  \end{subfigure}
  \hfill
  \begin{subfigure}[]{0.45\textwidth}
    \centering
    \includegraphics[width=\textwidth]{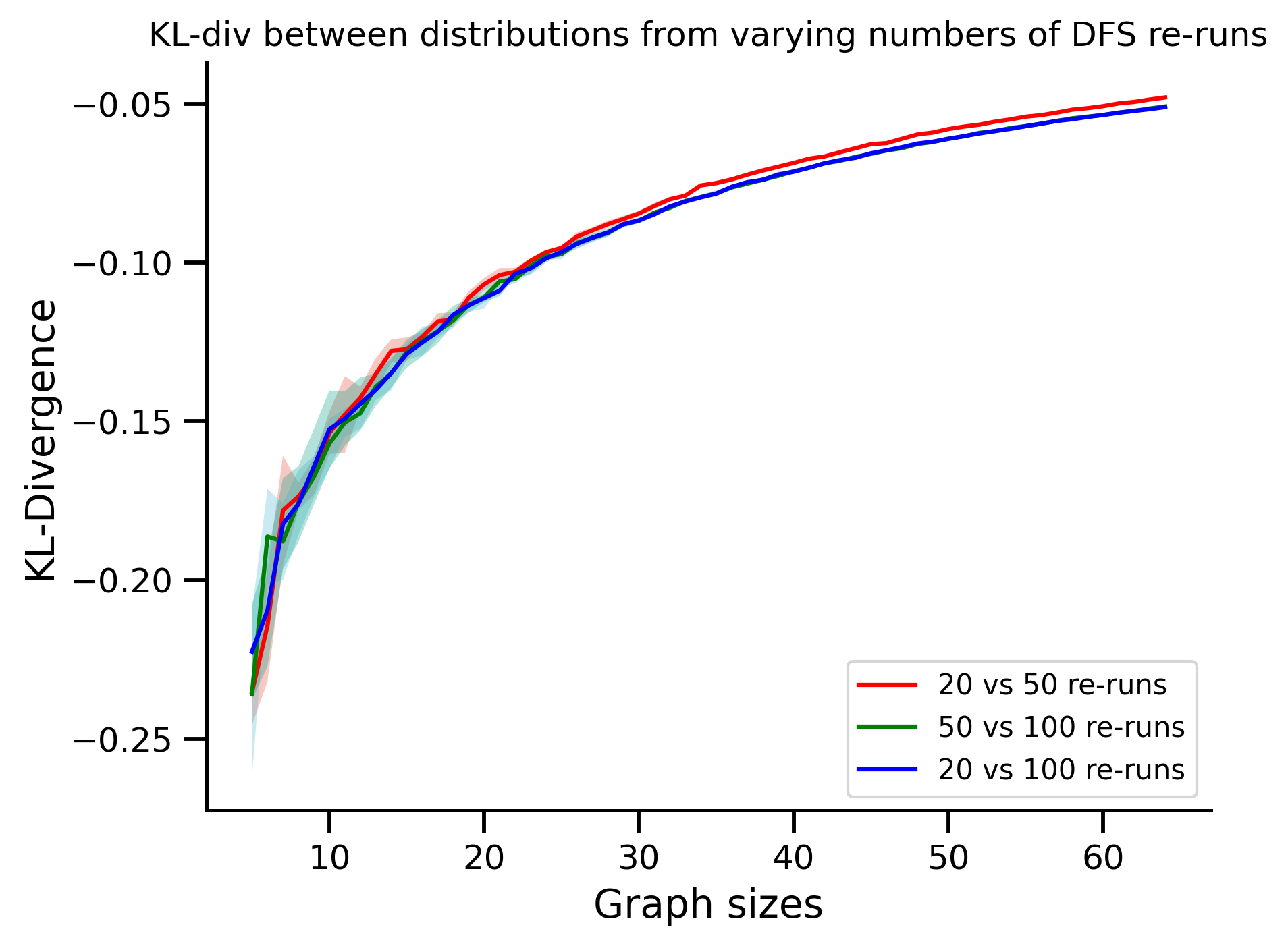}
    \caption{Depth-first search}
    \label{fig:bf_sampling_n64}
  \end{subfigure}
  \caption{Pairwise mean distribution differences for DFS and BF for 20, 50, and 100 reruns for graphs of sizes 5 through 64.}
  \label{fig:num_reruns}
\end{figure}

When creating our training data by generating distributions over parents for DFS and Bellman-Ford, we need to consider how often we run the randomised versions of the deterministic algorithms on a graph in order to obtain a good distribution over solutions. To this end, we consider the graph sizes $n \in \{5, \dots, 64\}$ and generate 100 graphs for each size and run an algorithm (DFS or BF) 20, 50, and 100 times for each graph. For each graph, we then consider the KL-Divergence between each pair of distributions generated by a different number of algorithm re-runs (we compare the distribution generated from 20 runs to the distribution generated from 50 runs and to the one generated from 100 runs, et cetera). We plot the mean KL-Divergence along with its standard deviation for each pairwise comparison between re-run numbers for all considered sizes. In Figure \ref{fig:num_reruns}, we see that while there are differences between the distributions generated from a varying number of algorithm re-runs, the distribution differences do not change markedly when going from 20 to 100 versus from 50 to 100 re-runs. Therefore, by increasing the number of re-runs, it appears that we do not gain enough information to justify the added computational cost of an increased re-run number. Therefore, we settle on 20 as a reasonable value for the number of algorithm re-runs.

%\newpage
\section{Validating the model output distributions}\label{sec: Appendix: validating distributions}
Going beyond our analysis of the validity of solutions obtained from the method presented in this paper, we investigate the coverage of the space of all solutions (i.e. the space of possible solutions we can obtain from executing an algorithm on a given graph $G$) of our methods. Using DFS as an example, we compare the average number of solutions when using DFS on graphs of a given size to the average number of unique and valid solutions when using our sampling method on the outputs of our model on the test data generated from those same graphs. This affords insight into the proportion of the space of possible solutions our method is capable of covering, which is particularly relevant to applications of Neural Algorithmic Reasoning in which using many different solutions is important, such as in cyberphysical systems. 

\subsection{Validating model output distributions for Bellman-Ford}\label{sec: Appendix: validating distributions, BF}
\begin{figure}[H]
  \centering
  \begin{subfigure}[]{0.44\textwidth}
    \centering
    \includegraphics[width=\textwidth]{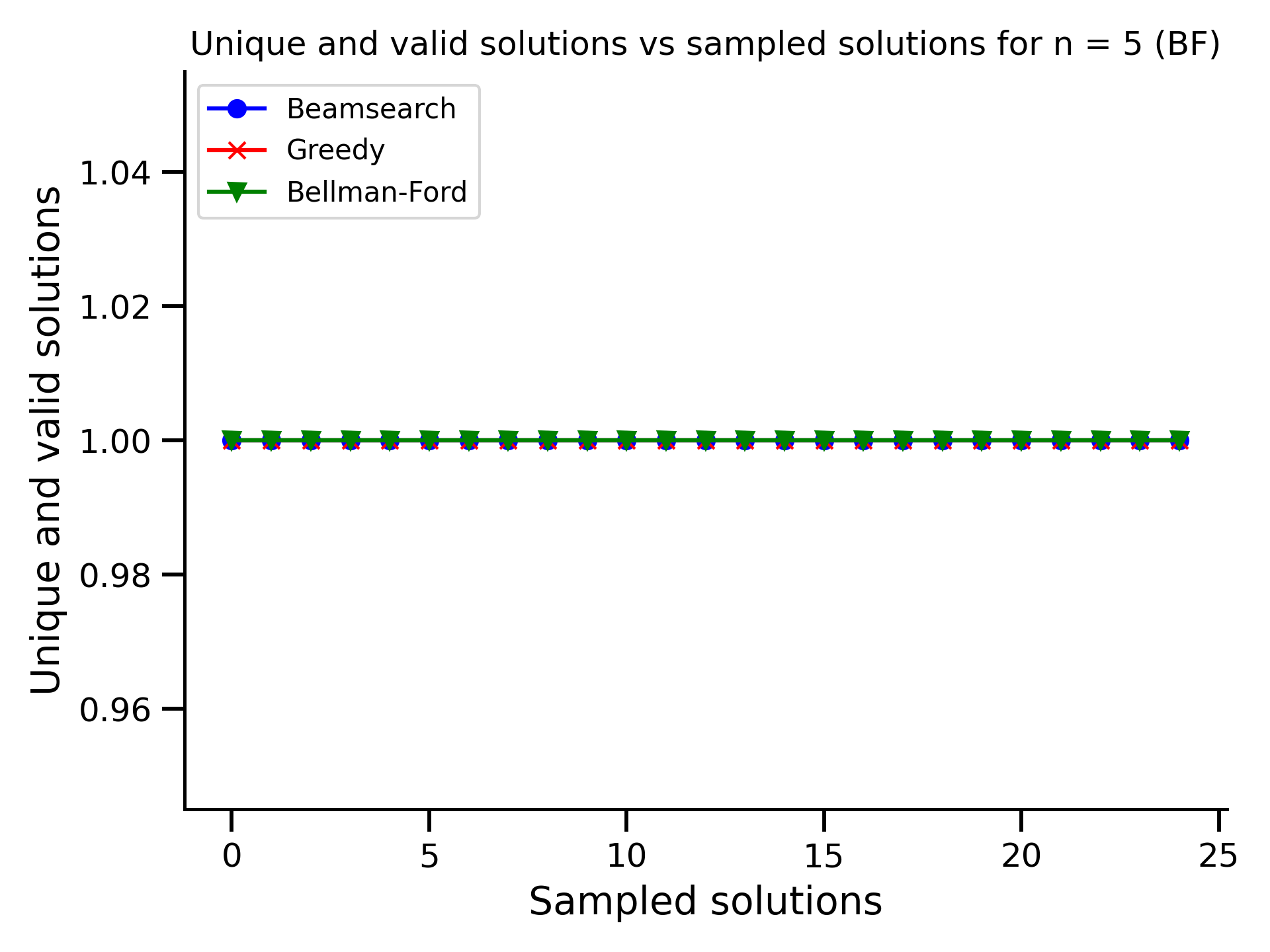}
    \caption{5-vertex Graphs.}
    \label{fig:plot_unique_by_extracted_5}
  \end{subfigure}
  \hfill
  \begin{subfigure}[]{0.44\textwidth}
    \centering
    \includegraphics[width=\textwidth]{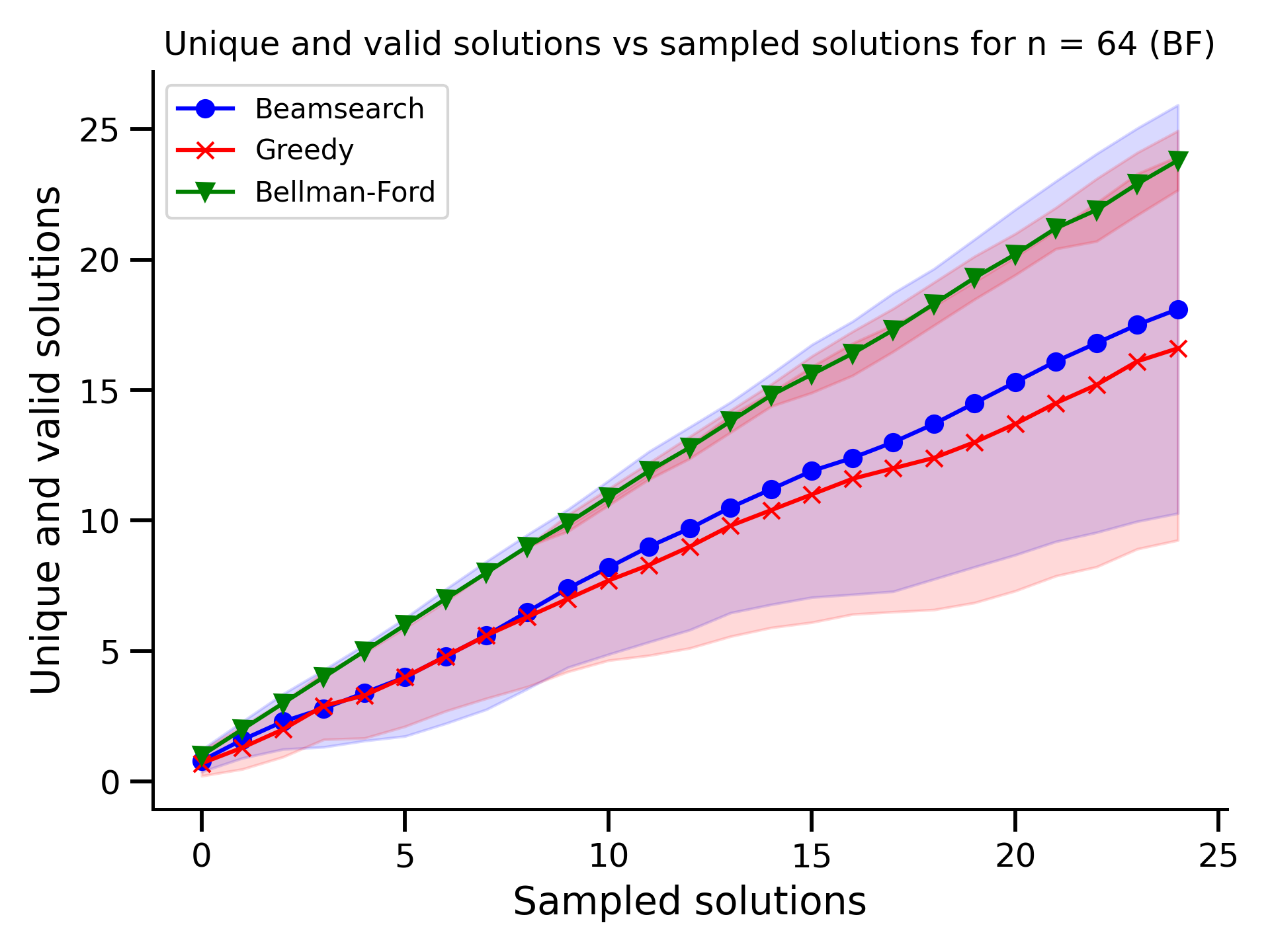}
    \caption{64-vertex Graphs.}
    \label{fig:plot_unique_by_extracted_64}
  \end{subfigure}
  \caption{The number of valid unique solutions found by our sampling methods on model output distributions and Bellman-Ford on the original graphs for 10 graphs and 25 samples.}
  \label{fig:edge_reuse_line_bf}
\end{figure}
Figure \ref{fig:plot_unique_by_extracted_5} shows that our sampling methods simply find the one valid solutions that exists for graphs of size 5 in all cases, which validates both our model and the sampling methods. In Figure \ref{fig:plot_unique_by_extracted_64}, we can see that even for out-of-distribution graphs, the model output distributions are good enough for our sampling methods to return a high diversity of valid solutions, even within the margin error of straightforwardly applying BF to the input graph. This coverage of the solution space is very encouraging for further research into NAR with multiple solutions, as it shows the potential advantage of using NAR to find all possible solutions, which is relevant to many applications, as discussed in the body of the paper.

\begin{figure}[]
  \centering
  \begin{subfigure}{\linewidth}
    \centering
    \includegraphics[width=0.4\linewidth, keepaspectratio]{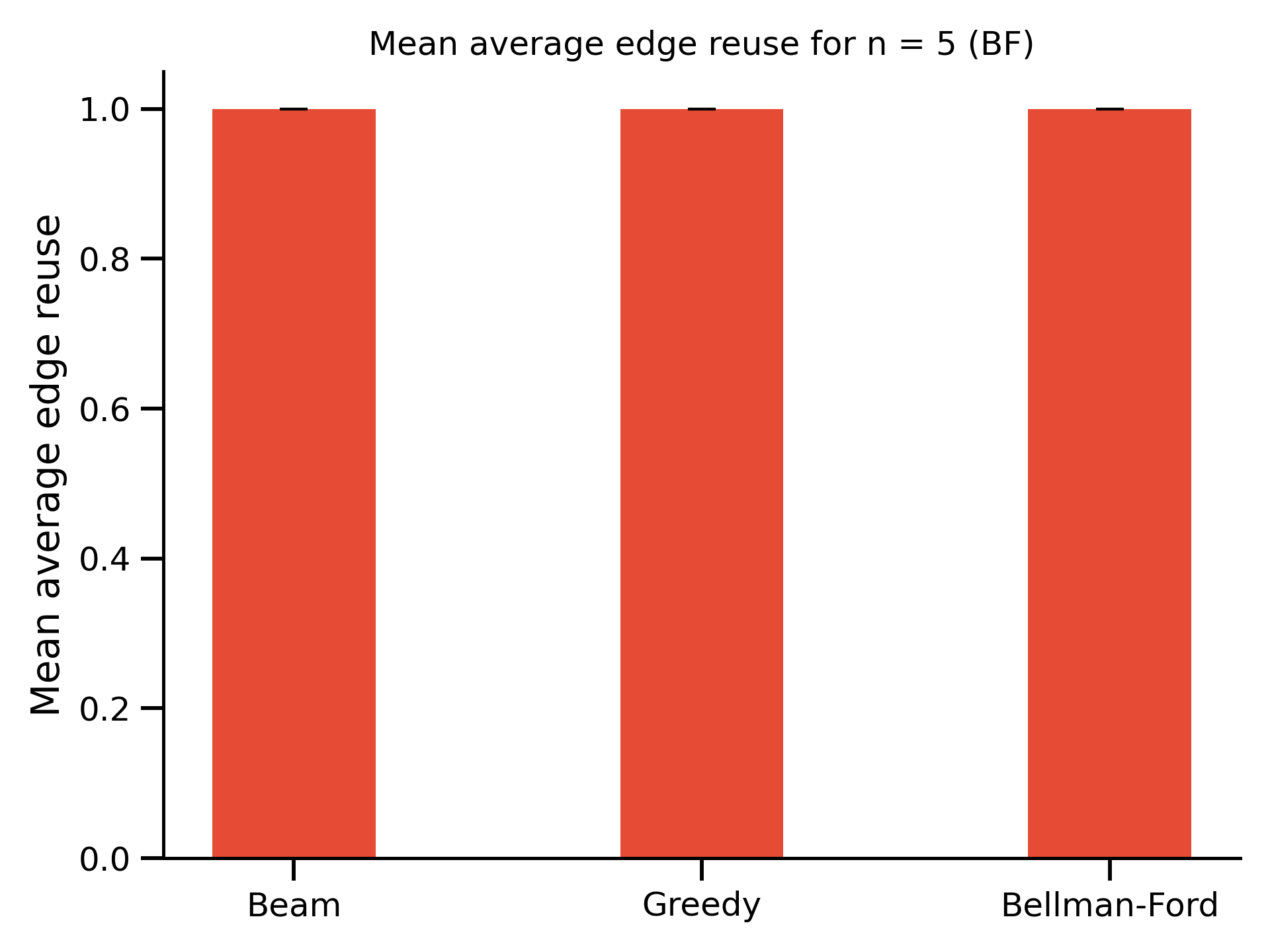} % replace with your image
    \caption{5-vertex Graphs.}
\label{fig:edge_reuse_mean_5_bf}
  \end{subfigure}
  \vspace{1em}
  \begin{subfigure}{\linewidth}
    \centering
\includegraphics[width=0.4\linewidth, keepaspectratio]{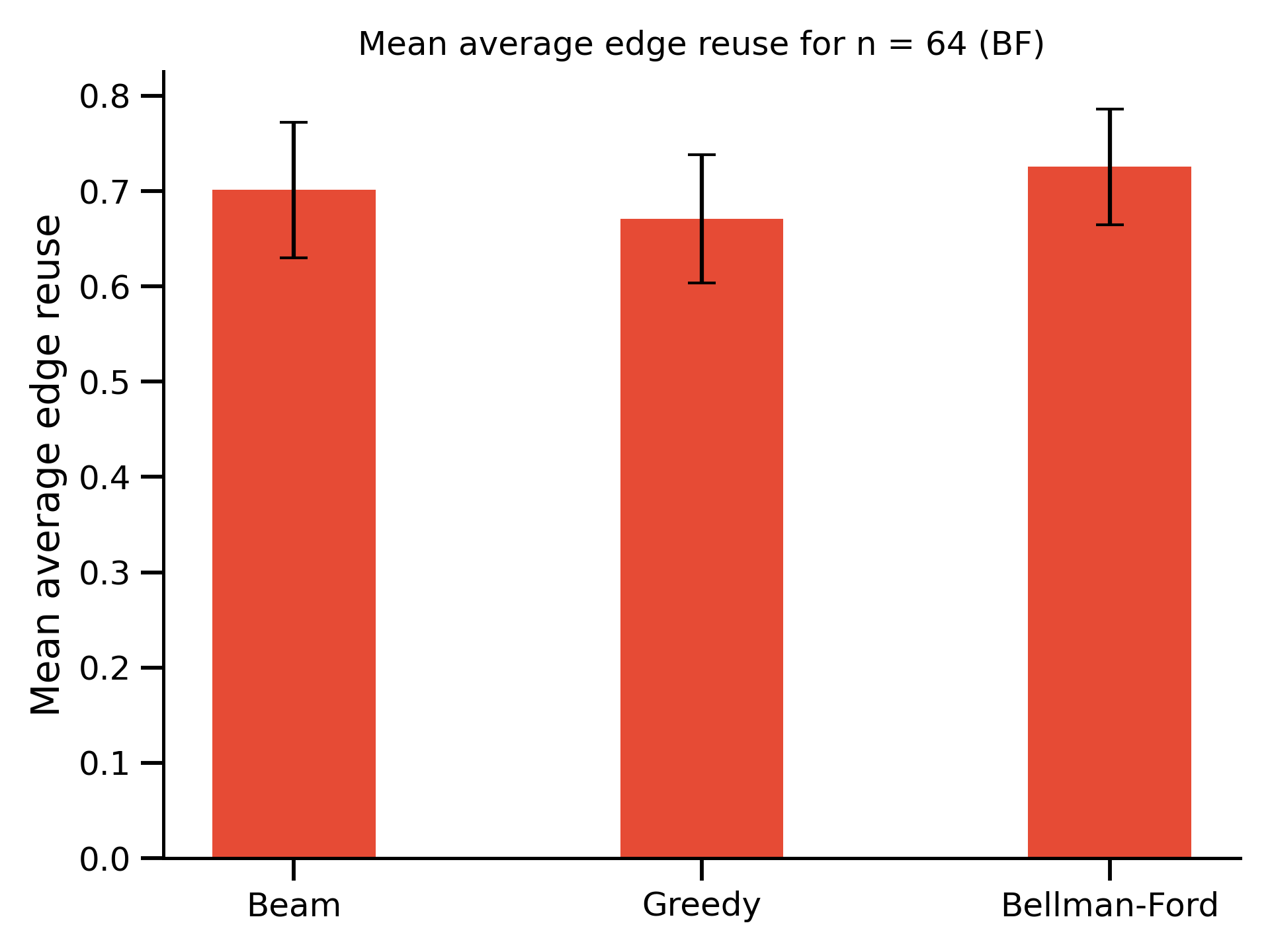} % replace with your image
    \caption{64-vertex Graphs.}
    \label{fig:edge_reuse_mean_64_bf}
  \end{subfigure}
    \caption{Mean edge reuse for Bellman-Ford sampling.}
\label{fig:edge_reuse_mean_nx_bf}
\end{figure}

\begin{figure}[H]
  \begin{subfigure}[]{0.45\textwidth}
    \centering
    \includegraphics[width=\textwidth]{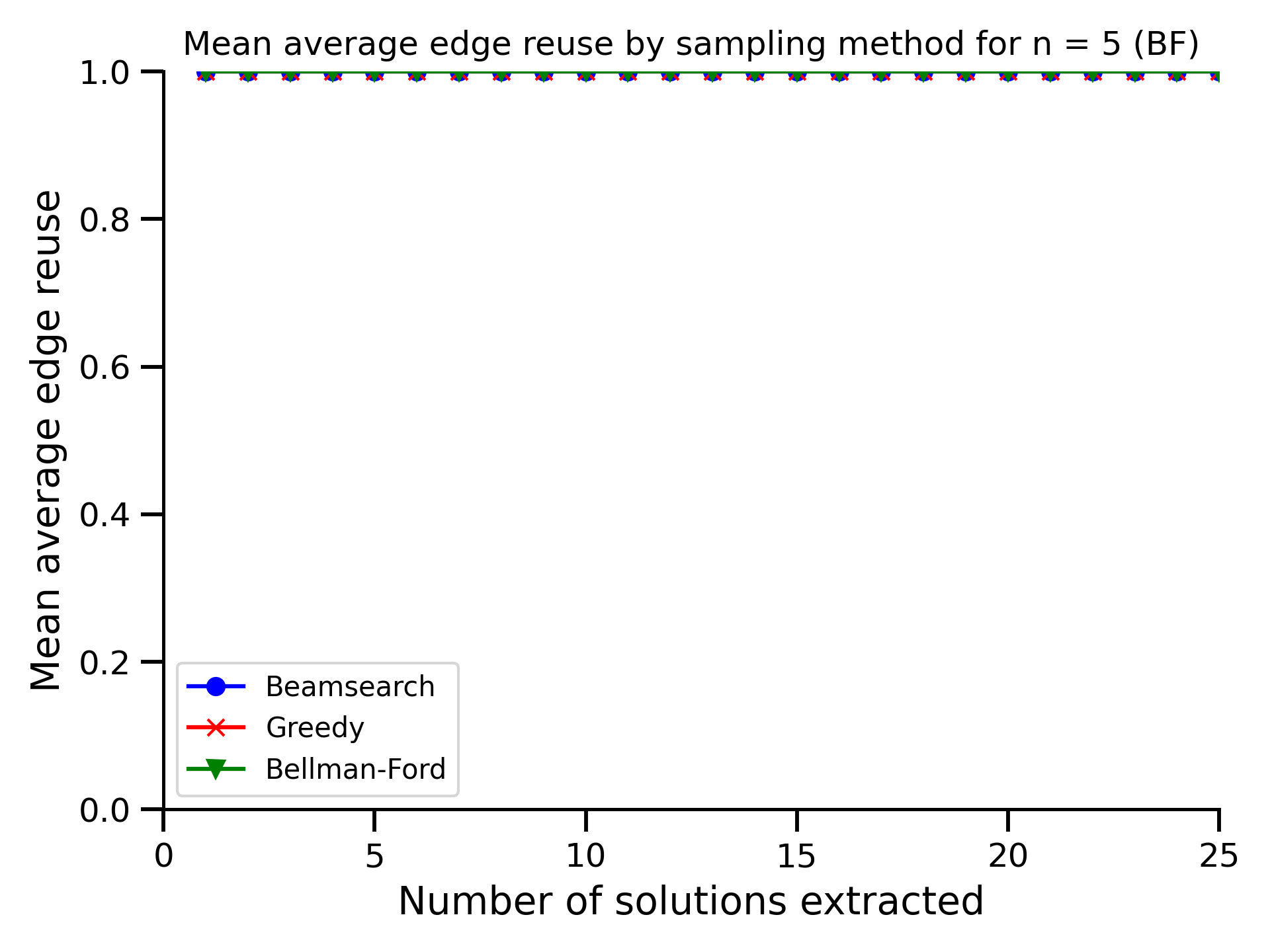}
    \caption{5-vertex Graphs.}
    \label{fig:fig:edge_reuse_line64_bf}
  \end{subfigure}
  \hfill
  \begin{subfigure}[]{0.45\textwidth}
    \centering
    \includegraphics[width=\textwidth]{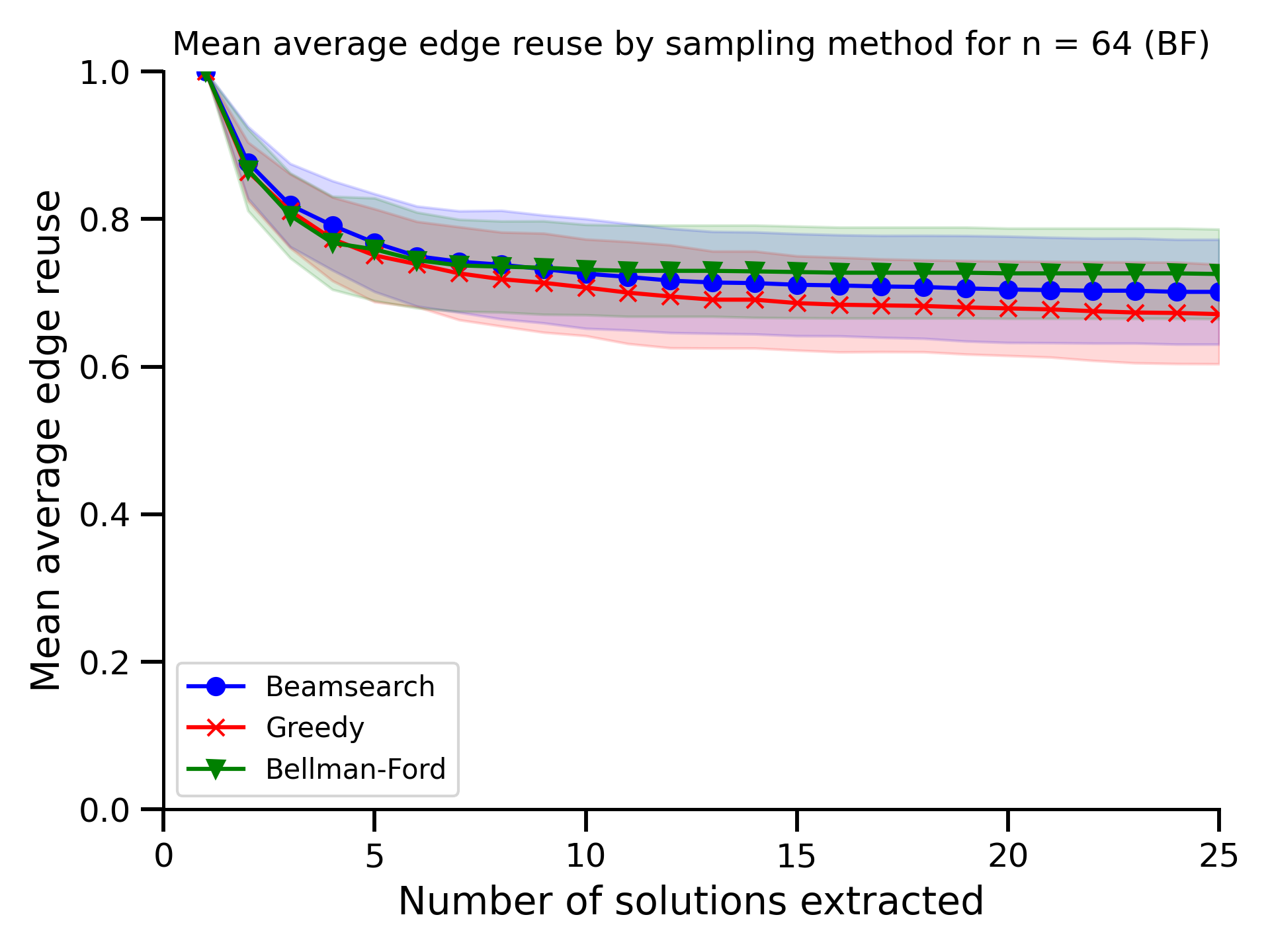}
    \caption{64-vertex Graphs.}
    \label{fig:edge_reuse_line64_bf}
  \end{subfigure}
  \caption{Evolution of the mean edge reuse for different Bellman-Ford sampling methods.}
  \label{fig:edge_reuse_line_bf_nx}
\end{figure}

Figures \ref{fig:edge_reuse_mean_nx_bf} and \ref{fig:edge_reuse_line_bf_nx} show that for both considered graph sizes, the proportions of edges in common among sampled solutions is similar across sampling methods (referred to further as \textit{mean edge reuse}), which is unsurprising given the nature of single-source shortest paths, where only a certain subset of edges is relevant to finding the shortest paths. 

\subsection{Validating model output distributions for DFS}
\begin{figure}[ht]
    \includegraphics[width=0.45\textwidth]{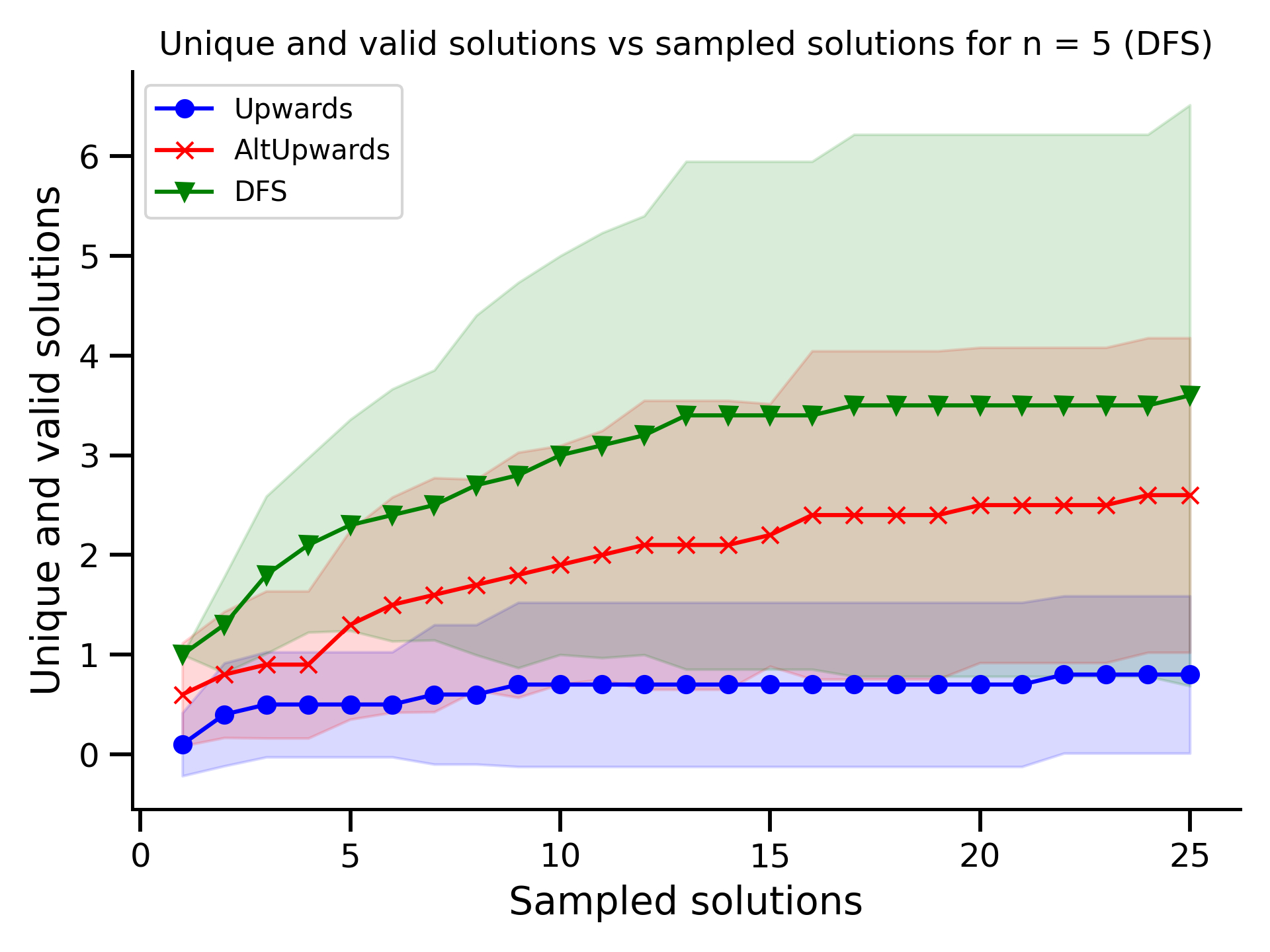}
  \caption{The number of valid unique solutions found by our sampling methods on model output distributions and DFS on the original graphs for 10 graphs and 25 samples.}
  \label{fig:plot_unique_by_extracted_5_dfs}
\end{figure}

Figure \ref{fig:plot_unique_by_extracted_5_dfs} shows the marked advantage of \textit{AltUpwards} over \textit{Upwards}, as \textit{AltUpwards} finds as many solutions as DFS on small graphs, where \textit{Upwards} struggles to return any diversity of solutions. However, we see in our results that our methods do not generalise to larger graph sizes, possibly indicating that the chosen encoding of the multiple solutions as probability distributions over predecessors is not suitable for problems with a very diverse set of solutions for a given graph.
% \begin{figure}[H]
%   \centering
%   \begin{subfigure}[]{0.45\textwidth}
%     \centering
%     \includegraphics[width=\textwidth]{valid_dist_figs/edge_reuse_mean_5_dfs.png}
%     \caption{5-vertex Graphs.}
%     \label{fig:edge_reuse_mean_5_dfs}
%   \end{subfigure}
%   \label{fig:edge_reuse_mean_nx_dfs}
% \end{figure}
% For those solutions that are valid, however, Figures \ref{fig:edge_reuse_mean_nx_dfs} and \ref{fig:edge_reuse_lineplot_nx_dfs} show that our sampling methods yield solutions that are similarly diverse (as measured by the mean proportion of edges they share between them) as running DFS on the input graph. 
%\newpage
\section{Training Details} \label{sec: model details}
\subsection{Model architecture and parameters}
To give a good baseline, we use the default CLRS Benchmark NAR neural network \cite{clrs-benchmark}. %We have an encoder, a processor network, and a decoder.
Our processor network is an MPNN \cite{mpnns} with a two standard NAR adjustments: namely, gating and triplet reasoning \cite{ibarz2022generalist}. Our encoder is standard: a linear layer initialised with xavier on scalars. Our decoder is 4 linear layers — three with hidden dimension (128) and a final layer with dimension one. The network is recurrent.   

%The network is recurrent, with a number of steps determined by (one can measure by the number of hint-probes pushed, although we refrain from using hints for simplicity here. HELP IM COUNTING PROBING.PUSHES AND GETTING 280 for 20 SOLUTIONS IN DFS SIZE 4 GRAPHS. IS IT REALLY 14 STEPS? OR 280 STEPS? WHATS A STEP?

%To focus on methodological feasibility, We use the most default NAR network possible

% _Type.DOBRIK_AND_DANILO: decoders = (linear(hidden_dim), linear(hidden_dim), linear(hidden_dim), linear(1))

% We use the CLRS benchmark default triplet\_gmpnn as processor neural network, with custom encoders and decoders adapted to handle our modified input-output pairs for BF and DFS.

% Our NN is a gated message passing neural network with triplet reasoning.
% In general, we use the CLRS default triplet\_gmpnn hyperparameters and implementation \cite{ibarz2022generalist}, with the deviations of our approach reported alongside standard parameters in Table \ref{tab: trainingparams}.
% One could by changing a flag. 
% We refrain from reporting multiple architectures in the interest of exhaustive analysis of a generic MPNN, as a good baseline for , to show more methodological ...

\begin{table}[]
    \centering
    \begin{tabular}{cc}
         Parameter & Value \\
         \toprule Processor & triplet\_gmpnn \\
         \midrule Encoder & xavier\_on\_scalars\\
         \midrule Decoder & 4 linear layers \\
         \midrule \#Parameters & 391225 \\
%         \midrule \#Layers & \\
         \midrule \#Hidden Units & 128 \\
         \midrule Learning Rate & 0.001 \\
         \midrule Batch Size & 1\\
         \midrule Training Graph Sizes & 4, 7, 11, 13, 16 \\
         \midrule Training Set Size & 1000 \\
         \midrule Testing Graph Sizes & 5, 64\\
         \midrule Testing Set Size & 10 \\
         \bottomrule
    \end{tabular}
    \caption{Training parameters}
    \label{tab: trainingparams}
\end{table}

%processor_factory = triplet_gmpnn
%processor = MPNN(
          %out_size=out_size,
          %msgs_mlp_sizes=[out_size, out_size],
          %use_ln=use_ln,
          %use_triplets=True,
          %nb_triplet_fts=nb_triplet_fts,
          %gated=True,
% in file processors.py line 420haha I love u

% {'processor_factory': <function get_processor_factory.<locals>._factory at 0x169c663a0>, 'hidden_dim': 128, 'encode_hints': False, 'decode_hints': False, 'encoder_init': 'xavier_on_scalars', 'use_lstm': False, 'learning_rate': 0.001, 'grad_clip_max_norm': 1.0, 'checkpoint_path': '/tmp/CLRS30', 'freeze_processor': False, 'dropout_prob': 0.0, 'hint_teacher_forcing': 0.0, 'hint_repred_mode': 'soft', 'nb_msg_passing_steps': 1} NOTE nb_msg... is number msg passing steps per hint, which doesnt matter since no hints

    % elif kind == 'gmpnn':
    %   processor = MPNN(
    %       out_size=out_size,
    %       msgs_mlp_sizes=[out_size, out_size],
    %       use_ln=use_ln,
    %       use_triplets=False,
    %       nb_triplet_fts=nb_triplet_fts,
    %       gated=True,
    %   )

\subsection{Time characteristics}
We run our experiments using a publicly available compute provider that uses NVIDIA P100 GPUs. Including evaluation of our model every 50 steps, training the NARs for 10,000 steps is very time-inexpensive. We summarize the training times in Table \ref{tab: traintime}.

\begin{table}[]
    \centering
    \begin{tabular}{c|cc}
         & $n = 5$ & $n = 64$ \\\midrule
         BF & 240sec.& 248sec. \\
         DFS &531sec. & 550sec. \\
        \bottomrule
    \end{tabular}
    \caption{Training times for the conducted experiments in seconds}
    \label{tab: traintime}
\end{table}
\section{The Graph Accuracy Metric}\label{sec: Appendix Graph Accuracy}
With multiple solutions, node level accuracy is poorly defined. Ordinarily in a single-solution framework, a node is counted as accurately predicted if its prediction matches the true solution: If the true solution is $[0,1]$ and the reasoner predicts $[1,1]$, its scores 50\% on node accuracy. Put another way, node accuracy scores the fraction of a candidate solution that matches the label solution. In a multiple solution framework, that metric becomes untenable. First, when there are multiple true solutions, producing all the solutions to give partial credit is computationally expensive. Second and more importantly, if $[0,1]$ and $[1,0]$ are both true solutions, node accuracy scores $[1,1]$ at 100\% even if it is incorrect. As a result, we use `graph` accuracy, as defined in \cite{minder2023salsaclrssparsescalablebenchmark}. If the model predicts $[1,1]$, and it is not a correct solution, the solution is scored with 0\% graph accuracy, regardless of whether $[0,1]$ or $[1,0]$ are true solutions. Formally, graph accuracy for multiple solutions scores a solution $S$ as correct for a graph $G$ and algorithm $A$ if and only if $S \gets A(G)$ for some run of $A$. To give a sense of relative strictness, we provide node level and graph level accuracies for the CLRS benchmark trained for 10,000 steps in Table \ref{tab: working vanilla accuracy}. Interpreting node accuracy as the probability that a NNs prediction is correct for any node, we would expect $graph_{accuracy} = (node_{accuracy})^{n}$. We can see that the discrepancy between graph accuracy and node accuracy increases with graph size.

\begin{table}[h!]
	\centering
\caption{Node and Graph Accuracy for Bellman-Ford and DFS with regular CLRS training. Network trained for 10,000 steps. Predictions on 64 graphs of each size. Rounded averages over 5 runs.} \label{tab: working vanilla accuracy}
\begin{subtable}{\linewidth}
\centering
\begin{tabular}{l@{\hskip 3pt}c@{\hskip 3pt}c@{\hskip 3pt}c@{\hskip 3pt}c}
    & \multicolumn{2}{c}{Bellman Ford}  \\
		\cmidrule{2-3} 
        
& Node & Graph \\
		\cmidrule{1-3}
(n=4) 
&$1.00 \pm 0.00$
&$1.00 \pm 0.14$
\\
		
(n=16)
&$0.99 \pm 0.00$
&$0.91 \pm 0.05$
\\

(n=64) 
&$0.98 \pm 0.00$
&$0.21 \pm 0.08$
\\ \cmidrule{0-2}
\end{tabular}
\end{subtable}
\vspace{1em}
\begin{subtable}{\linewidth}
\centering
\begin{tabular}{l@{\hskip 3pt}c@{\hskip 3pt}c@{\hskip 3pt}c@{\hskip 3pt}c}
  \multirow{2.5}{*}{} &  \multicolumn{2}{c}{DFS}  \\
		\cmidrule{2-3}
        
& Node & Graph \\
    \cmidrule{1-3}
(n=4) 
&$1.00 \pm 0.00$
&$1.00 \pm 0.00$
\\
		
(n=16)
&$0.91 \pm 0.02$
&$0.41 \pm 0.11$
\\

(n=64) 
&$0.39 \pm 0.01$
&$0.00 \pm 0.00$
\\
\cmidrule{0-2}
\end{tabular}
\end{subtable}
\end{table}

\section{Permuting Inputs}
We explain our experiments conducted by permuting inputs, a fundamentally different approach than predicting solution distributions. A graph has $n!$ ways of isomorphically labeling the nodes. One might hope that feeding a NAR the same graph with two different node labellings would produce distinct solutions. The results are largely negative. Regardless of what one permutes, NARs rarely predict more than one solution, and when they do, the additional solution is usually incorrect.

\subsection{Permuting All Inputs}

The following experiments generate a permutation $\sigma$, and permute each item consistently. So, $\sigma(A_{ij}) = A_{\sigma(i),\sigma(j)}$ for all $i,j$. 

\subsubsection{Bellman Ford}

For Bellman Ford, CLRS NARs predict on 
\begin{itemize}
    \item A, a weighted adjacency matrix
    \item s, a one-hot vector encoding the start node
    \item adj, a copy of $A$ without self loops and with all edge weights set to 1.0
    \item pos, a vector of floats (useful for other algorithms, fairly irrelevant here)
\end{itemize}

%Most importantly, we evaluate whether a predicted solution $P$ is valid for the original adjacency matrix $ogA$ with start node $S$, which we call `Valid(ogA,S,P)'.  
%Note `Valid(ogA,S,P)'sense in which this method might usefully produce multiple solutions, because original adjacency matrix and start node. 
We report several metrics of BF validity (Table \ref{tab: w bf permute everything}). 
The variables `ogA, ogS, ogP' stand for the original adjacency matrix, the original start node, and the NN's prediction on the original graph. 
We obtain `ogP' by applying the inverse permutation to P, the NN's actual prediction.
The variables `A, S' represent the permuted adjacency matrix and the permuted start node, which are fed to the NN to produce prediction `P'. Put succinctly, we have $P \gets NN(A,S,adj,pos)$.
Table \ref{tab: w bf permute everything} shows that Bellman Ford Prediction validity is the same on permuted and original graphs, as one might expect since shortest paths are invariant to node relabeling. 

We also report several metrics of diversity. Diversity in this context refers to whether different solutions involve different node labels in each position: two isomorphic solutions may be counted as different. Second, `Distinctness' tests whether solutions are non-isomorphic, by comparing `ogP'. Third, `VD' measures the fraction of valid solutions that are distinct. In Table \ref{tab: w bf permute everything}, we see that the network only ever predicts one valid solution on 5 permutations of the same input data.

%For context, we also collect the stats `Valid(A,S,P)' for whether $S$ is valid on the permuted adjacency matrix $A$, which the NN took as input.
%We thirdly report whether the inverse permutation of $S$, $ogS$ represents the solution and is used to distinguish isomorphic but relabeled solutions.
%Unpermuting $S$, we see that the NN really predicts the same solution on each of the isomorphic graphs.

\begin{table}[h!]
	\centering
\caption{Bellman Ford mean accuracies and proportions of distinct solutions (0 to 1) when permuting all inputs; 5 permutations, with standard deviations over 5 runs.}\label{tab: w bf permute everything}

\begin{tabular}{l@{\hskip 3pt}c@{\hskip 3pt}c@{\hskip 3pt}c@{\hskip 3pt}c}
    & \multicolumn{3}{c}{Validity}  \\
		\cmidrule(lr){2-4} 
        
& valid(ogA, ogS, ogP) & valid(A,S,P) & valid(ogA,S,P)\\
		\midrule
(n=4) 
&$1.00 \pm 0.00$
&$1.00 \pm 0.00$
&$0.32 \pm 0.06$
\\
		
(n=16)
&$0.91 \pm 0.05$
&$0.91 \pm 0.05$
&$0.00 \pm 0.00$
\\

(n=64) 
&$0.21 \pm 0.08$
&$0.21 \pm 0.08$
&$0.00 \pm 000$
\\ \hline
\\
  \multirow{2.5}{*}{} & \multicolumn{3}{c}{Variety}  \\
		\cmidrule(lr){2-4} 
		& Diversity & Distinctness & VD  \\
		\midrule
(n=4)
&$0.61 \pm 0.09$
&$0.20 \pm 0.00$
&$0.21 \pm 0.03$
\\

(n=16)
&$0.99 \pm 0.01$
&$0.20 \pm 0.00$
&$0.20 \pm 0.00$
\\

(n=64)
&$1.00 \pm 0.00$
&$0.20 \pm 0.00$
&$0.00 \pm 0.00$
\\
		\bottomrule
\end{tabular}
\end{table}

\subsubsection{DFS}

\begin{table}[]
	\centering
\caption{DFS mean accuracies and proportions of distinct solutions (0 to 1) when permuting all inputs; 5 permutations, with standard deviations over 5 runs.}\label{tab: w dfs permute everything}
\begin{subtable}{\linewidth}
        \centering
\begin{tabular}{l@{\hskip 3pt}c@{\hskip 3pt}c@{\hskip 3pt}c@{\hskip 3pt}c}
    & \multicolumn{4}{c}{Validity}  \\
		\cmidrule(lr){2-5} 
        
& H(ogA, ogP) & H(A,P) & H(ogA,P) & J(ogA,P)\\
		\midrule
(n=4) 
&$1.00 \pm 0.00$
&$0.19 \pm 0.14$
&$0.08 \pm 0.04$
&$0.28 \pm 0.05$
\\
		
(n=16)
&$0.48 \pm 0.12$
&$0.05 \pm 0.05$
&$0.01 \pm 0.01$
&$0.05 \pm 0.04$
\\

(n=64) 
&$0.00 \pm 0.00$
&$0.00 \pm 0.00$
&$0.00 \pm 0.00$
&$0.00 \pm 0.00$
\\ \hline
\end{tabular}
   \end{subtable}

    \vspace{1em}
\begin{subtable}{\linewidth}
        \centering
\begin{tabular}{l@{\hskip 3pt}c@{\hskip 3pt}c@{\hskip 3pt}c@{\hskip 3pt}c}
  \multirow{2.5}{*}{} & \multicolumn{3}{c}{Variety}  \\
		\cmidrule{2-4}
		& Variety & Distinctness & VD  \\
		\cmidrule{1-4}
(n=4)
&$0.93 \pm 0.05$
&$0.20 \pm 0.00$
&$0.38 \pm 0.23$
\\

(n=16)
&$1.00 \pm 0.01$
&$0.20 \pm 0.00$
&$0.27 \pm 0.11$
\\

(n=64)
&$1.00 \pm 0.00$
&$0.20 \pm 0.00$
&$0.00 \pm 0.00$
\\
\cmidrule{1-4}
\end{tabular}
\end{subtable}
\end{table}

We report several metrics of DFS validity (Table \ref{tab: w dfs permute everything}). 
As with Bellman Ford, variables `ogA, ogP' stand for the original adjacency matrix, the original start node, and the NN's prediction on the original graph. 
The variable `A` represents the permuted adjacency matrix fed to the NN to predict `P`. We thus have $$P \gets NN(A,adj,pos).$$

For DFS, the CLRS benchmark baseline models \cite{clrs-benchmark} predict solutions from
\begin{itemize}
    \item A, a weighted adjacency matrix
    \item adj, a copy of $A$ without self loops and with all edge weights set to 1.0
    \item pos, a vector of floats (useful for other algorithms, fairly irrelevant here)
\end{itemize}

We have two different algorithms for verification: `H' is algorithm \ref{Alg:Henry}, and determines whether P is producible by RDFSO. 
The second algorithm `J' is a relaxed version of `H', determining if P is producible by RDFS with restarts in any order.
Table \ref{tab: w dfs permute everything} shows that RDFSO prediction validity is hurt by permutation, as one might expect since a correct solution can become incorrect after relabeling (e.g. by starting at any node other than 0). 
Table \ref{tab: w dfs permute everything} distinctness also shows that the NN predicts the same isomorphic solution each time, adapting it to the relabeling. Unfortunately, permuting all the inputs does not yield multiple correct solutions.

\subsection{Only Permuting Pos}

Both Bellman-Ford and DFS networks take an input $pos$. 
One can think of $pos$ as a proxy for node indices (the floats are randomly-spaced but always in ascending order: $pos[0]<pos[1]<...pos[n]$. 
One might therefore wonder if training a network to tiebreak according to $pos$ might produce better results.
However, when we train and predict only permuting $pos$, the models still do not produce multiple correct solutions. 
For Bellman Ford, the network predicts only one solution (Table \ref{tab: mt bf permute smth}).
For DFS, accuracy is too low to see one correct solution for n=16 graphs, let alone multiple for n=64 (Table \ref{tab: mt dfs permute smth}).
A possible reason for this failure is that situations where tiebreaking is necessary are rare, so tiebreaking according to $pos$ may be more difficult to learn than tiebreaking according to index.
This hypothesis is supported by the lower accuracies in  Table \ref{tab: mt vanilla} than Table \ref{tab: working vanilla accuracy}, despite the same training for the same model architectures: the only difference is Table \ref{tab: mt vanilla} labels tiebreak according to a permuted $pos$, whereas Table \ref{tab: working vanilla accuracy} labels tiebreak according to node index.
\begin{table}[]
	\centering
\caption{Bellman-Ford mean accuracies and proportions of distinct solutions (0 to 1) when permuting the $pos$ input; 5 permutations, standard deviations over 5 runs.}\label{tab: mt bf permute smth}
\begin{subtable}{\linewidth}
\centering
\begin{tabular}{l@{\hskip 3pt}c@{\hskip 3pt}c@{\hskip 3pt}c@{\hskip 3pt}c}
    & \multicolumn{3}{c}{Validity}  \\
		\cmidrule(lr){2-4} 
        
& valid(ogA, ogS, ogP) & valid(A,S,P) & valid(ogA,S,P)\\
		\midrule
(n=4) 
&$1.00 \pm 0.00$
&$1.00 \pm 0.00$
&$0.33 \pm 0.10$
\\
		
(n=16)
&$0.86 \pm 0.03$
&$0.86 \pm 0.03$
&$0.00 \pm 0.00$
\\

(n=64) 
&$0.19 \pm 0.03$
&$0.19 \pm 0.03$
&$0.00 \pm 000$
\\ \hline
\end{tabular}
   \end{subtable}

    \vspace{1em}
    \begin{subtable}{\linewidth}
    \centering
    \begin{tabular}{l@{\hskip 3pt}c@{\hskip 3pt}c@{\hskip 3pt}c}
\multicolumn{3}{c}{Variety}  \\
		\cmidrule{2-3} 
		& Distinctness & VD  \\
		\cmidrule{1-3} 
(n=4)
&$0.20 \pm 0.09$
&$0.20 \pm 0.00$
\\

(n=16)
&$0.24 \pm 0.01$
&$0.20 \pm 0.00$
\\

(n=64)
&$0.75 \pm 0.04$
&$0.20 \pm 0.00$
\\
		\cmidrule{1-3} 
\end{tabular}
\end{subtable}
\end{table}
% \subsubsection{DFS}
\begin{table}[]
	\centering
\caption{DFS mean accuracies and proportions of distinct solutions (0 to 1) when permuting the $pos$ input; 5 permutations, standard deviations over 5 runs.}\label{tab: mt dfs permute smth}
\begin{subtable}{\linewidth}
        \centering
\begin{tabular}{l@{\hskip 3pt}c@{\hskip 3pt}c@{\hskip 3pt}c@{\hskip 3pt}c}
    & \multicolumn{4}{c}{Validity}  \\
		\cmidrule(lr){2-5} 
        
& H(ogA, ogP) & H(A,P) & H(ogA,P) & J(ogA,P)\\
		\midrule
(n=4) 
&$0.06 \pm 0.01$
&$0.09 \pm 0.04$
&$0.01 \pm 0.01$
&$0.04 \pm 0.01$
\\
		
(n=16)
&$0.00 \pm 0.00$
&$0.00 \pm 0.00$
&$0.00 \pm 0.00$
&$0.00 \pm 0.00$
\\

(n=64) 
&$0.00 \pm 0.00$
&$0.00 \pm 0.00$
&$0.00 \pm 0.00$
&$0.00 \pm 0.00$
\end{tabular}
    \end{subtable}

    \vspace{1em}
\begin{subtable}{\linewidth}
\centering
\begin{tabular}{l@{\hskip 3pt}c@{\hskip 3pt}c@{\hskip 3pt}c@{\hskip 3pt}c}
\multicolumn{3}{c}{Variety}  \\
\cmidrule(lr){1-3} 
		& Distinctness & VD  \\
		\cmidrule{1-3} 
(n=4)
&$0.45 \pm 0.05$
&$0.28 \pm 0.14$
\\

(n=16)
&$1.00 \pm 0.00$
&$0.00 \pm 0.00$
\\

(n=64)
&$1.00 \pm 0.00$
&$0.00 \pm 0.00$
\\
		\cmidrule{1-3} 
\end{tabular}
\end{subtable}
\end{table}
\begin{table}[]
	\centering
\caption{Node and Graph Accuracy for Bellman-Ford and DFS with CLRS training where algorithms tiebreak in $pos$ order. Network trained for ten thousand steps. Predicting on 64 graphs of each size. Standard deviations over 5 runs}\label{tab: mt vanilla}
\begin{subtable}{\linewidth}
\centering
\begin{tabular}{l@{\hskip 3pt}c@{\hskip 3pt}c@{\hskip 3pt}}
    & \multicolumn{2}{c}{Bellman-Ford}  \\
		\cmidrule{2-3} 
        
& Node & Graph \\
		\cmidrule{1-3}
(n=4) 
&$1.00 \pm 0.00$
&$1.00 \pm 0.00$
\\
		
(n=16)
&$0.99 \pm 0.00$
&$0.88 \pm 0.03$
\\

(n=64) 
&$0.97 \pm 0.00$
&$0.14 \pm 0.06$
\\ \cmidrule{0-2}
\end{tabular}
\end{subtable}
\vspace{1em}
\begin{subtable}{\linewidth}
\centering
\begin{tabular}{l@{\hskip 3pt}c@{\hskip 3pt}c@{\hskip 3pt}}
&\multicolumn{2}{c}{Depth-First Search}  \\
		\cmidrule{2-3}
        
& Node & Graph \\
    \cmidrule{1-3}
(n=4) 
&$0.58 \pm 0.04$
&$0.06 \pm 0.01$
\\
		
(n=16)
&$0.13 \pm 0.01$
&$0.00 \pm 0.00$
\\

(n=64) 
&$0.03 \pm 0.01$
&$0.06 \pm 0.00$
\\
\cmidrule{0-2}
\end{tabular}
\end{subtable}
\end{table}
\end{document}